%% file: main.tex
\documentclass[runningheads,a4paper,11pt,twoside]{llncs}
\usepackage{Packages}

\begin{document}

%
\title{On the Generalization of PINNs outside the training domain and the Hyperparameters influencing it}


%
\titlerunning{}
%
\author{
    Andrea Bonfanti$^{1,2,3}$, Roberto Santana$^2$, Marco Ellero$^{3,4,5}$, Babak Gholami$^1$
}
\authorrunning{A. Bonfanti, R. Santana, M. Ellero, B. Gholami}
%
\institute{$^{(1)}$BMW Group, Digital Campus Munich, Munich, Germany;\\
$^{(2)}$University of the Basque Country (UPV/EHU), San Sebastian–Donostia, Spain;\\
$^{(3)}$BCAM - Basque Center for Applied Mathematics, Bilbao, Spain;\\
$^{(4)}$IKERBASQUE, Basque Foundation for Science, Bilbao, Spain;\\
$^{(5)}$Zienkiewicz Centre for Computational Engineering (ZCCE), Swansea University, Swansea, UK.
}%
\maketitle
\begin{abstract}
\input{Sections/000_Abstract}

\end{abstract}

\section{Introduction}\label{sec:050_Introduction}
\input{Sections/050_Introduction}

\section{Physics-Informed Neural Networks and Generalization}\label{sec:100_Method}
\input{Sections/100_Method}

\section{Problem Description and Explorative Analysis}\label{sec:X00_Numerics}
\input{Sections/X00_Numerics.tex}

\section{Hyperparameter study through the Generalization Level}\label{sec:200_Results}
\input{Sections/200_Results}

\section{Discussion on the effect of the algorithmic setup}\label{sec:300_Discussion}
\input{Sections/300_Discussion}

\section{Conclusion}\label{sec:X_Conclusion}\input{Sections/X_Conclusion}





\end{document}

%% file: Sections/000_Abstract.tex
Physics-Informed Neural Networks (PINNs) are Neural Network architectures trained to emulate solutions of differential equations without the necessity of solution data. They are currently ubiquitous in the scientific literature due to their flexible and promising settings. However, very little of the available research provides practical studies that aim for a better quantitative understanding of such architecture and its functioning. In this paper, we perform an empirical analysis of the behavior of PINN predictions outside their training domain. The primary goal is to investigate the scenarios in which a PINN can provide consistent predictions outside the training area. Thereinafter, we assess whether the algorithmic setup of PINNs can influence their potential for generalization and showcase the respective effect on the prediction. The results obtained in this study returns insightful and at times counterintuitive perspectives which can be highly relevant for architectures which combines PINNs with domain decomposition and/or adaptive training strategies.


%% file: Sections/050_Introduction.tex

Physics-Informed Neural Networks (PINNs) were first introduced in the 90s \cite{veryfirstPINN, Lagaris_1998} as an alternative method to solve ordinary and partial differential equations, based on the structure and approximation capabilities of Neural Networks. In late 2017, Raissi et al. reintroduced such architecture with the aforementioned name in two publications \cite{raissi2017physicsI, raissi2017physicsII}, aimed at investigating its range of applicability to PDE-based forward and inverse problems. The analysis provided in those papers shows that PINNs are capable of solving classical monodimensional and bidimensional PDEs with little training effort and with satisfying precision. However, the complexity of the training routine rapidly increases if the underlying equations become more complex or the domain of the solution is large and/or multidimensional.

There are several reasons behind the choice of a Machine Learning-based approach, such as PINNs, instead of a classical numerical method to solve a PDE. A PINN architecture represents a simple and meshless solution that does not require high-level knowledge of the underlying equations and allows noisy data to be incorporated into the training process, enabling the analysis of an incomplete set of data or partially understood phenomena \cite{yang2021bayesian}. On the other hand, especially due to their recent introduction in the field, PINNs lack the convergence guarantees that are characteristic of well-founded numerical solvers. Moreover, training a PINN can often be slow and sometimes unreliable for relatively complex equations in three or more dimensions \cite{pitfallsandfrustration}. That is, there is no certainty that a trained PINN reaches the correct solution, and sometimes including data in the training process might be necessary to obtain a reasonable output \cite{art:raissi2018hidden}. Despite their drawbacks, PINNs have already seen their applications for a large variety of fields ranging from biology \cite{PINNcardiovascularmri, zapf2022investigating}, meteorology\cite{PINNmetheo}, and system safety \cite{PINNsafety}. They are ubiquitous in recent research literature in general, due to the endless potential fields of applications. We refer the reader to \cite{PINNreview} for a more detailed overview.


In this paper, we shed light on the concept of generalization to points outside of the training domain for PINNs and hereinafter evaluate their potential in extrapolation with respect to the training hyperparameters of the network architecture. Generalization has a particular connotation for PINNs, since the models are aiming to predict a value which satisfies some deterministic properties and laws. Therefore, the concept of generalization of PINNs has to be carefully investigated through methods that minimize the effect of the stochasticity intrinsic to the training of machine learning models. For this purpose, we define a specific metric to quantify the generalizability of a PINN based on the algorithmic setup adopted.

Our paper is structured as follows: In Section \ref{sec:100_Method}, we provide a formal definition of the PINN architecture and define our new generalization error metric after clarifying the concept of generalization for PINNs in terms of the adopted training dataset. Then, in Section \ref{sec:X00_Numerics}, we introduce the test case adopted for the rest of the paper and perform a preliminary numerical and statistical analysis on the PINN architecture to explore its potential and limitations in terms of generalization. Section \ref{sec:200_Results} is then devoted to illustrating the results obtained in terms of the previously defined metric and the results are finally discussed in detail in Section \ref{sec:300_Discussion}. The paper concludes with Section \ref{sec:X_Conclusion}, which wraps up the main insights obtained through the experiments and provides recommendations for related future research directions.

\subsection{Related work and Contributions}

The analysis carried out in this paper is related to the quality of generalization of neural networks, which consists of the capacity of a learning machine to evaluate correctly data samples that are not close to those included in the training process. Generalization is a key property of machine learning algorithms as it indicates how well a trained model learns the input-output mapping for the considered dataset. In recent years, the concept of generalization for neural networks has gained increasing interest in the research community due to its atypical behavior in the overparametrized regime given by the double descent curve \cite{advani2020high, olson2018modern, mei2022generalization}. Despite the lack of available results in this field, such study is not easily applicable to the case of PINN architectures. Indeed, the goal of PINN is to learn a deterministic function that must satisfy some prescribed laws and is unique for the majority of the problems. Henceforth, applying classical probabilistic bounds does not provide exact information on the accuracy of a PINN, which is an essential result for any kind of engineering application. To the best of our knowledge, the only applicable bounds are found in the work of Mishra et al. \cite{mishra2022estimates} which focuses on deriving error estimates of trained PINN architectures, based on the value of their loss function and the number of sample points included in the training domain. However, the only bound applicable to the outside of the training domain grows exponentially with respect to the distance from the training boundaries. Intuitively, this suggests that the error of a PINN can not be guaranteed to be small when evaluated outside of the training area.

The error of a PINN outside of its training domain has not been explored systematically in the literature yet. Indeed, since no knowledge of points outside the training domain is given to the PINN architecture during training, it is natural to assume that a valid prediction cannot be guaranteed. The majority of the works in the literature that presented PINN predictions outside of their training domain were focused on domain decomposition, as for the case of $\textit{hp}$-VPINNs \cite{kharazmi2021hp} and Finite-Basis-PINNs  \cite{moseley2021finite}. In both publications, the paradigm of training a PINN to solve a PDE in its domain is enhanced by the usage of domain decomposition. Therefore, the prediction outside the training area is actually not used for predictions, but rather showcased to the reader for visualization purposes. However, it is important to be mindful of the behavior of PINN prediction outside the training area, specially when the decomposition relies on soft decompositions \cite{hu2022augmented} or learnable gating mechanisms \cite{bischof2022moepinn}. Indeed, since domain decomposition methods have become more and more used as an enhancement for PINN-based models, it is important to understand whether the prediction of the submodels can present diverging behaviour outside their training area, as that might introduce detrimental errors in the overall prediction and instability in the training proceedure.

Since it is not possible to provide analytical bounds for errors in predictions far from the training domain of a PINN, our goal is to numerically study the degradation of the predictions of a PINN outside of the training domain. The goal is then to understand whether the error in prediction can be bounded or controlled through architectural hyperparameters of the neural network employed. In line with the approach of structural analysis followed in this paper, we can find similar works as, for instance, \cite{autopinn} and \cite{exploringpinns}, which provide an analysis of the performance of PINNs based on several hyperparameters of the architecture, or \cite{adaptivesamplingPINNs}, which studies the effect of different sampling strategies when choosing the training data for PINNs. Another similar project is given by \cite{sharma2022accelerated}, where a new methodology proposed to train PINNs is analyzed through the lenses of the depth and width of the neural network used.

The analysis we provide in this paper differs from previous works in the following aspects: 

\begin{itemize}
    \item To the best of our knowledge, this is the first work to systematically study the behavior of PINNs outside of their training area;
    \item Statistical methods and tests are used to quantify and validate the significance of the impact of the hyperparameters of the PINN on the prediction of the network;
    \item An extensive investigation is made on the effect in generalization of varying the aforementioned hyperparameters of the PINN architecture and the effects obtained is discussed based on recent results from the scientific literature;
    \item A new generalization metric is constructed to be only weakly affected by stochasticity and based on the intrinsic properties of the structure of a neural network.
\end{itemize}

%% file: Sections/100_Method.tex
This section introduces the theoretical background on which this paper is based. First, we provide a brief definition of a PINN and highlight the aspects of this neural network that are relevant to our study. Thereinafter, we focus on the concept of generalization for a PINN: we explain why the classical statistical perspective does not apply to a PINN algorithm in a straightforward manner and introduce a possible method to study it.

\subsection{Physics-Informed Neural Networks}

The main concept behind the PINN architecture is to use a classical deep neural network to approximate the solution of a PDE in the domain where it is defined or where its values are needed. No a priori knowledge of the correct solution is necessary to train the model, except for its values at the boundaries of the domain and/or the initial condition of the PDE to make the problem well-posed. In the interior of the domain, where the solution is unknown, the network trains by minimizing the residuals of the underlying PDEs, computed through automatic differentiation. By doing so, the network tries to mimic a function that simultaneously fits the solution for the boundary and/or initial condition and is mathematically coherent with the underlying PDEs. Therefore, if the accuracy obtained after training is good enough, the neural network will become an approximation of the PDE solution in the domain of interest.

Formally, let us consider the problem of finding a function $u : \Omega_T \longrightarrow \mathbb{R}^n$ which solves a system of potentially non-linear partial differential equations, which we denote with $\mathcal{F}$, and which satisfies some initial and/or boundary conditions in some domain, which we will denote as $\Omega_T \subseteq \Omega \subset \mathbb{R}^d$. Finding the function $u$ amounts to solving the system of equations shown in Equation \eqref{ICBCproblem}.

\begin{equation}\label{ICBCproblem}
\begin{cases}
    \mathcal{F} [u(\vb{x}, t)] = 0 \qquad \quad & \forall (\vb{x}, t) \in \Omega_T \\
    u(\vb{x}, t) = u_b (\vb{x}, t) \qquad & \forall \vb{x}\in \partial \Omega_T \\
    u(\vb{x}, 0) = u_0 (\vb{x}) \qquad & \forall \vb{x}\in \Omega_T
    \end{cases}
\end{equation}

With $u_b$ and $u_0$ being respectively equal to the value of $u$ on $\partial \Omega_T$ and $\Omega_T \cap \{ (\vb{x},t) | t=0 \}$.
To solve the above system with a traditional PDE solver, the domain $\Omega_T$ has to be discretized through a mesh first, then the equation must be solved in discrete settings on the said mesh. Instead, $u$ can be approximated with a neural network $u_\theta : \Omega \longrightarrow \mathbb{R}^d$ of parameters $\theta \in \mathbb{R}^{N_\theta}$, with $N_\theta$ representing the number of learnable parameters. Therefore, the problem of solving the PDE becomes equivalent to finding the values of $\theta$ which make $u_\theta$ approximate $u$ with a prescribed accuracy in $\Omega_T$.
Finding those values requires the usage of a set of $N$ locations, named ``collocation points" $\{ (\vb{x}_i, t_i) \}_{i=1}^N$. Collocation points are typically randomly sampled points that belong to $\Omega_T$ and indicate where the residuals of the equations $\mathcal{F}$ are penalized, and a set of $N_b$ points in $\partial\Omega_T$ that represent the boundary and initial conditions satisfied by $u$.

The network $u_\theta$ is referred to as ``Physics-Informed" since its goal is to fit the values at the border while being skewed to minimize residuals of the equations represented by the functional $\mathcal{F}$. The typical loss function used for a Physics-Informed neural network is expressed in Equation \eqref{loss}.

\begin{equation}\label{loss}
\mathcal{L} (\theta, \{ (\vb{x}_i, t_i) \}_{i=1}^N, \{ (\vb{x}_j, t_j) \}_{j=1}^{N_b} )  =  \sum_{i=1}^{N}  ||\mathcal{F} [ u_\theta (\vb{x}_i, t_i)] || ^2  + \sum_{j=1}^{N_b} ||u_\theta (\vb{x}_j, t_j) - u (\vb{x}_j, t_j) || ^2
\end{equation}

It is clear from the definition of $\mathcal{L}$ that its minimizer is the correct solution of the PDE, assuming that the problem is well-posed.  However, learning the analytically exact solution with a neural network is practically cumbersome and theoretically infeasible due to limited machine precision. The solution $u_\theta$ obtained after training only aims to minimize its loss function in the points sampled before the training, included in some domain $\Omega_T \subseteq \Omega$. Once the training is done successfully, it is possible to accurately predict the value of $u$ in $\Omega_T$. This is possible thanks to the continuity of the target function and the known generalization power of neural networks.

\subsection{What is Generalization for a PINN and how to measure it?}\label{sec:what_gen_how_to}

The concept of generalization for an arbitrary machine learning architecture refers to its ability to provide accurate outputs when the input data is somewhat distant to the points used to train the model. Neural networks are widely known to present a natural tendency to generalize well to unseen data samples. This property of neural networks is what allows PINNs to learn the target PDE solution everywhere by using scattered points in the domain $\Omega_T$. The PINN does not have access to the true PDE solution at those points but, nevertheless, it can use them to learn the target function since it just needs to minimize the equation residuals. Those points can be interpreted as the dataset used by a PINN, and they roughly represent the domain $\Omega_T$ rather than the target PDE. 

Once the model is trained, one can evaluate the accuracy of the PINN with some new locations in $\Omega_T$. However, the accuracy obtained does not indicate the generalization potential of the model, since it is usually evaluated through new locations in $\Omega_T$ where the points used during training are dense. Therefore, a more suitable way to evaluate the generalization of a PINN can be given by studying the behaviour of its output outside of the convex hull defined by the training points.
Indeed, $u_\theta$ is trained until it accurately approximates $u$ inside $\Omega_T$ but there is no guarantee that the provided approximation is also valuable in $\Omega \setminus \Omega_T$, where $u$ evolves through its analytical extension. To study the generalization potential of PINNs, we define and use the value $G_l^\epsilon$, which we will refer to as ``Generalization Level" ($G_l$). The chosen metric is affected as little as possible by the stochasticity intrinsic to the training of a neural network. Indeed, we expect high GL values to indicate a more suitable structural choice for a PINN.

Consider the set of parameters of a PINN to be $\{ \theta_i \}_{i=1}^{N_\theta}  \in \Theta \subseteq \mathbb{R}^{N_\theta}$, and the remaining hyperparameters of the its training algorithm to be collected in an additional vector variable $H \in \mathcal{H}_{\text{PINN}}$. Consider also $u_\theta$ to identify a PINN initialized and trained according to the hyperparameters $H$ and with trainable parameters in $\Theta$. We define the GL of the hyperparameters of a PINN as in Definition \ref{generalization_metric}.

\begin{equation}\label{generalization_metric}
     G^\epsilon _l (\Theta, H ) = \min_{u_\theta} \Bigl\{ \max_{\Omega_G \subseteq \Omega \setminus \Omega_T} \bigl\{ \frac{l(\Omega_G)}{l(\Omega_T)} \quad \text{s.t.:} \\ || u(x) - u_\theta (x) || \leq \epsilon \quad \forall x\in \Omega_G  \bigr\} \Bigr\}
\end{equation}

Here, the function $l$ represents the area -- length for one-dimensional cases -- of the respective domain, while $\epsilon$ is a constant threshold defined by the underlying problem. Therefore, $G_l$ takes as input the space of the parameters $\Theta$ and the hyperparameters used for training a PINN and outputs the relative size of the biggest area outside the training domain in which the error remains bounded by $\epsilon$ for all the networks. This metric can be considered as the length of the $x$-segment outside of the training domain in which no significant variability in prediction can be visible and the prediction is accurate. The value of $G_l$ depends on the structure and dimensionality of $\Theta$ and on the values of $H\in \mathcal{H}_{\text{PINN}}$, which in this paper are considered as $H = (N_{\text{CP}}, A)$: the number of collocation points $N_{\text{CP}}$, and the size of the training domain $A$.\\

\subsubsection{Parameters influencing the potential in generalization}

It is reasonable to ask whether this extrapolation potential is affected by some of the hyperparameters of the neural network or the algorithmic setup of the PINN training. Among the possibly influential parameters, we consider the following ones:

\begin{itemize}
    \item \textit{Network complexity}: The number of neurons and the number of layers used for the neural network supposed to mimic the solution of the chosen PDE.
    \item \textit{Collocation points}: The number of reference locations in the PDE domain considered to compute the loss function of the PINN during training.
    \item \textit{Domain size}: The size of the domain $\Omega_T$ where the PINN is trained to solve the PDE.
\end{itemize}

There are also other components of the PINN architecture that can influence the behavior of the solution outside the training domain. The complexity of the target function is a clear example which is not treated in this paper as it represents an external component that we are not supposed to know a priori when training a PINN. Moreover, the kind of machine learning architecture used -- adding skip connections, dropout or a residual block --  can also affect the quality of the generalization obtained. We restrict our study to the base case of a multilayered perceptron for simplicity and due to the fact that the underlying problem chosen as a base case can be easily handled by such model.

%% file: Sections/X00_Numerics.tex
This section aims to provide numerical results to showcase and clarify the relevant features of a PINN architecture. Namely, we define the equation considered as a baseline for the upcoming study. Thereupon, we employ PINNs to solve said equation and use the trained architecture to highlight the main difference between PINNs and classical PDE solvers. Then, we use the same test case to clarify the issues of the generalization potential of a PINN, which is the core focus of this research.

For our scope, and without loss of generality, we choose a one-dimensional Poisson equation with a source term in the domain $[-\pi , \pi]$, presented in Equation \eqref{poisson}, as an academic example to showcase and analyze the output of a trained PINN. The presented equations are also solved through PINNs in \cite{lu2021deepxde} and such a choice for the equation relies on the richness of frequencies included in its solution, to have a target function that carries components with various frequencies. Moreover, training PINN architectures is also extremely efficient in one dimension, which facilitates the analysis provided in this paper.

\begin{equation}\label{poisson}
    \begin{cases}
      \frac{\partial ^2 u}{\partial x^2} = f \quad x \in [-\pi, \pi ]\\
      u(- \pi) = u(\pi) = 0
    \end{cases}\,.
\end{equation}

With $f: \mathbb{R}\longrightarrow \mathbb{R}$, the source term, being defined as:

\begin{equation}
    f(x) = \sum_{k=1}^5 2k \sin (2kx)
\end{equation}

The correct solution of this equation can be easily derived analytically by integrating twice the source term in $[-\pi, \pi]$ and setting the integration constants to zero, in order to match the desired values at the boundaries. The function resulting from this operation and solution to Equation \eqref{poisson} is given by:

\begin{equation}
    u(x) = \sum_{k=1}^5 \frac{1}{2k} \sin(2kx)
\end{equation}

\begin{figure} 
    \centering
  \subfloat[]{%
       \includegraphics[width=0.48\linewidth]{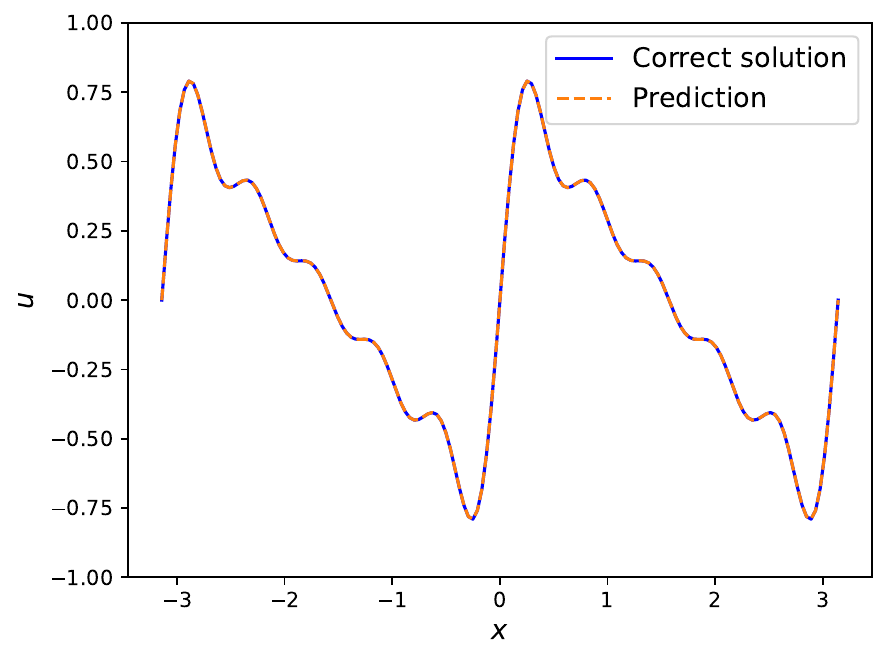}}
 \quad
  \subfloat[]{%
        \includegraphics[width=0.48\linewidth]{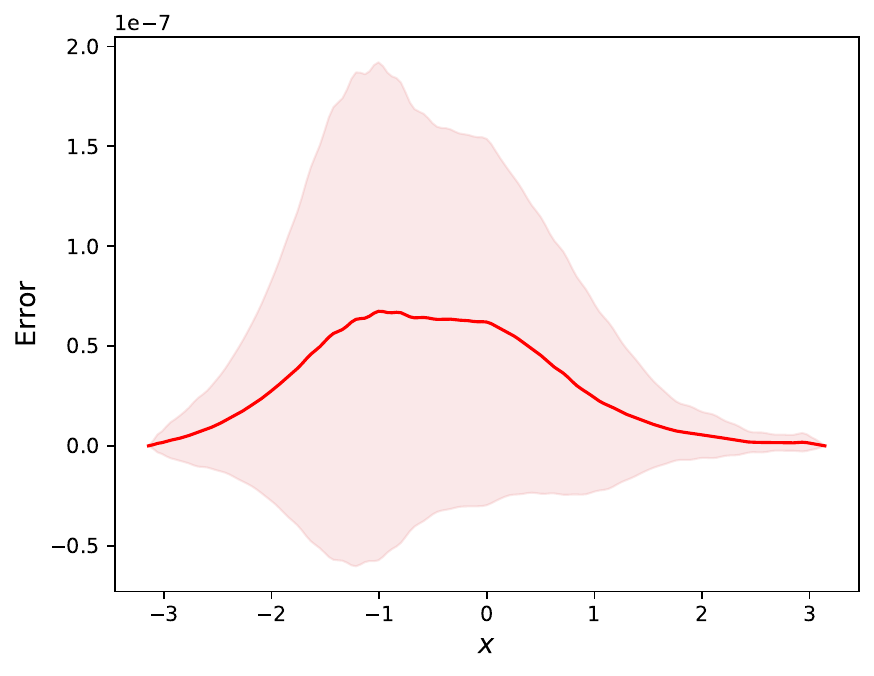}}

  \caption{Correct versus predicted solutions (a) and relative $L^2$ error (b) for 100 randomly initialized and trained PINN architectures. Variance is also included in both figures but it is too small to be visible in (a).}
  \label{PINN_Poisson_convergence} 
\end{figure}

To train the PINN architectures, we sample 100 collocation points in the domain $[-\pi, \pi]$ with the Latin hypercube sampling strategy and use a two-layers neural network architecture with 50 neurons per layer and hyperbolic tangent as activation function. To train the model, we restrict it to $10^4$ iterations of the Adam optimizer \cite{adamoptimizer}, followed by the L-BFGS optimizer \cite{lbfgsoptimizer} to fine-tune the results, limited to $10^4$ iterations and a tolerance of $10^{-8}$. The training process takes a few minutes to converge to a solution and outputs predictions that are visually indistinguishable from the correct solution, which can be seen in Figure \ref{PINN_Poisson_convergence}.

\begin{figure} 
    \centering
  \subfloat[]{%
       \includegraphics[width=0.33\linewidth]{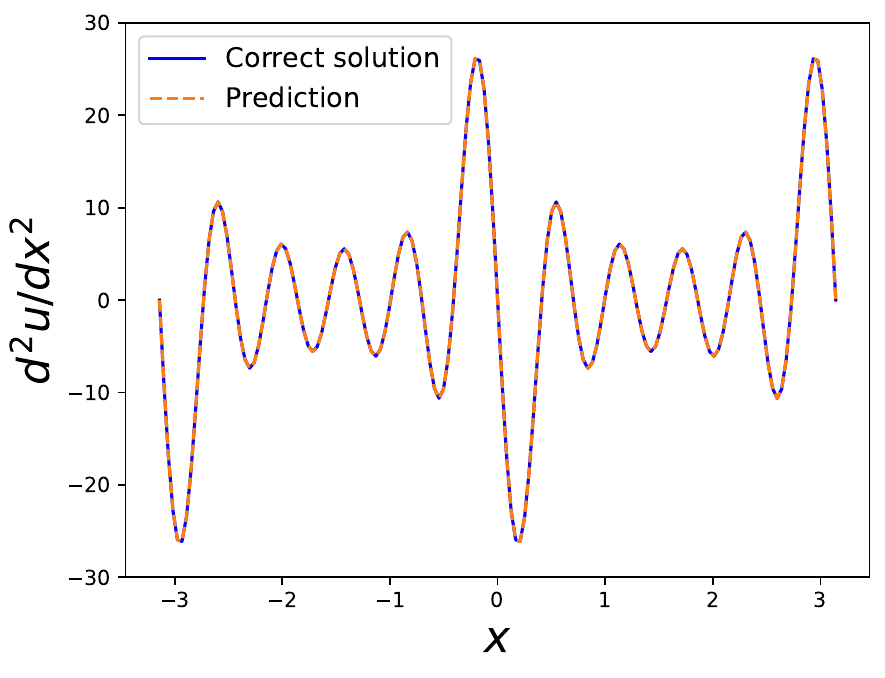}}
    \hfill
  \subfloat[]{%
        \includegraphics[width=0.33\linewidth]{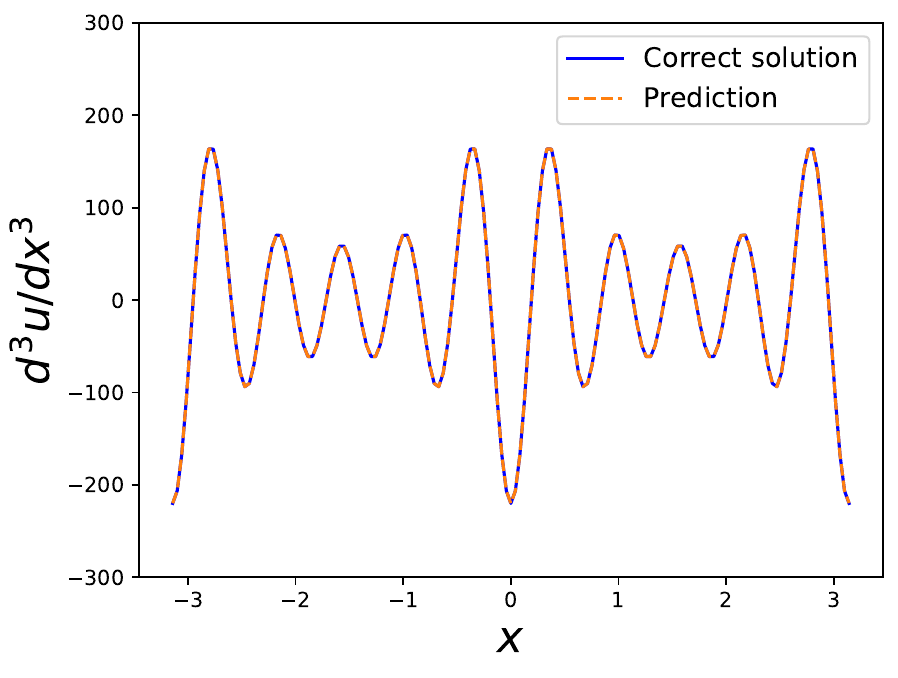}}
    \hfill
  \subfloat[]{%
        \includegraphics[width=0.33\linewidth]{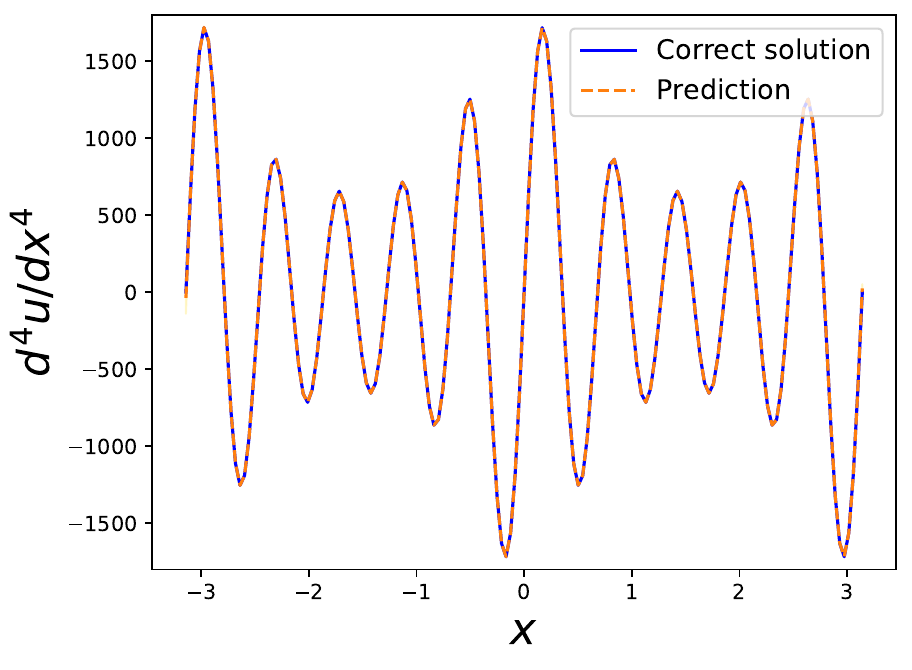}}
    \\
  \subfloat[]{%
        \includegraphics[width=0.33\linewidth]{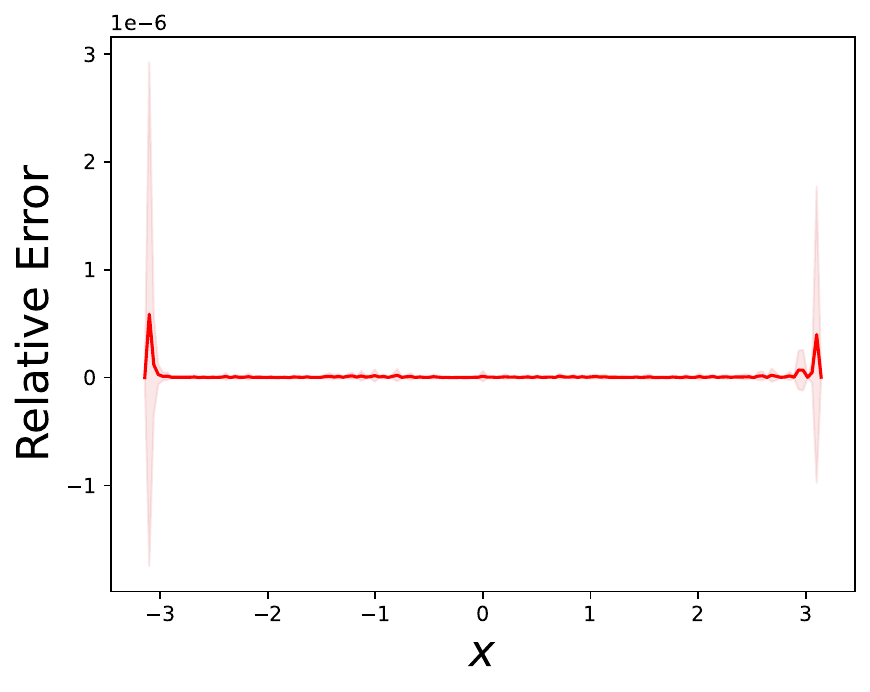}}
    \hfill
  \subfloat[]{%
        \includegraphics[width=0.33\linewidth]{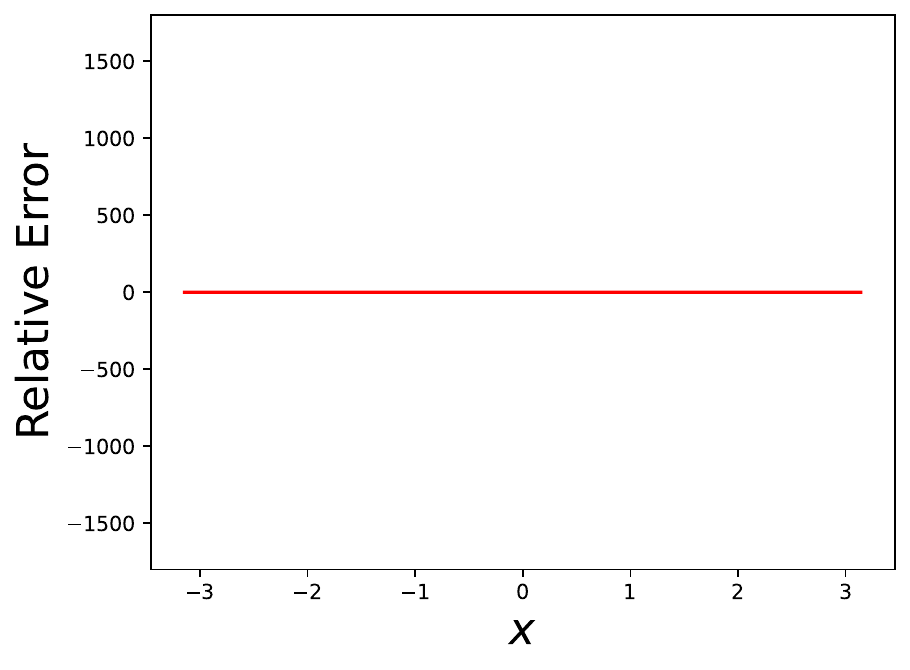}}
    \hfill
  \subfloat[]{%
        \includegraphics[width=0.33\linewidth]{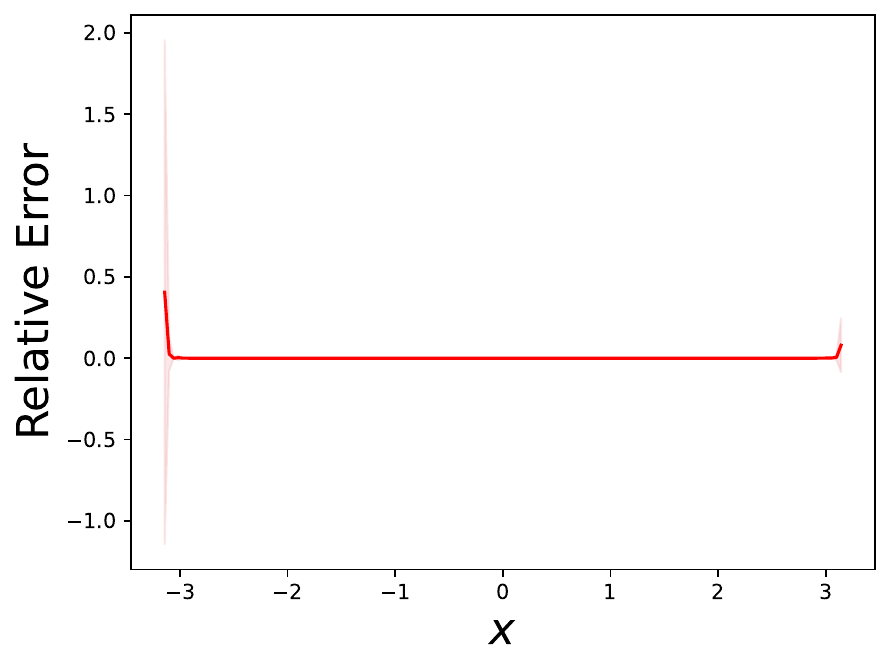}}
        
  \caption{True vs predicted values for the second, third and fourth derivative of the target solution for five randomly initialized and trained PINN architectures (respectively a,b, and c) along with their respective relative $L^2$ errors in prediction (d, e, and f). Variance of the prediction is also included but almost undistinguishable.}
  \label{PINN_Poisson_derivatives}
\end{figure}

In the same way, we used automatic differentiation to compute the equation residuals, we can also use it to access the derivatives of $u_\theta$ and compare them with the true ones. Due to the PDE that we are solving, it is already known that the prediction of the second derivative of $u$ will be accurate. Therefore, we compute and showcase the comparison of derivatives up to the fourth order along with the related relative $L^2$ error made in prediction. The choice of using the relative error rather than the classical squared error used in the previous section is due to the difference in magnitude of the values of $u$ and its first four derivatives, which goes from unitary order to $10^3$. The results are shown in Figure \ref{PINN_Poisson_derivatives}. Again, it is barely possible to distinguish the correct values from the ones obtained through the PINN architectures.

\subsection{Generalization beyond the Training Domain}

\begin{figure} 
    \centering
  \subfloat[]{%
       \includegraphics[width=0.33\linewidth]{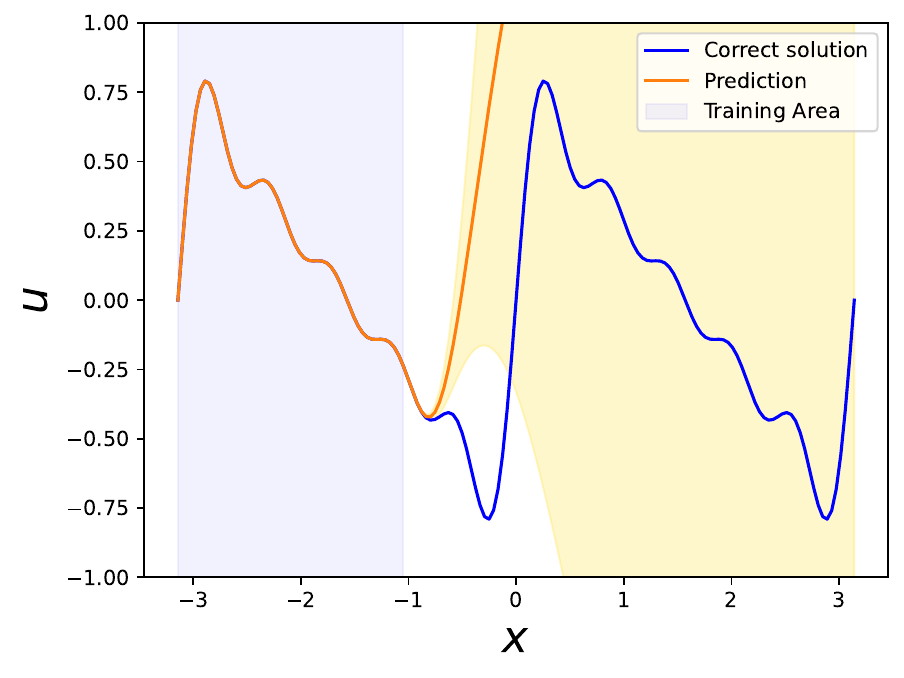}}
    \hfill
  \subfloat[]{%
        \includegraphics[width=0.33\linewidth]{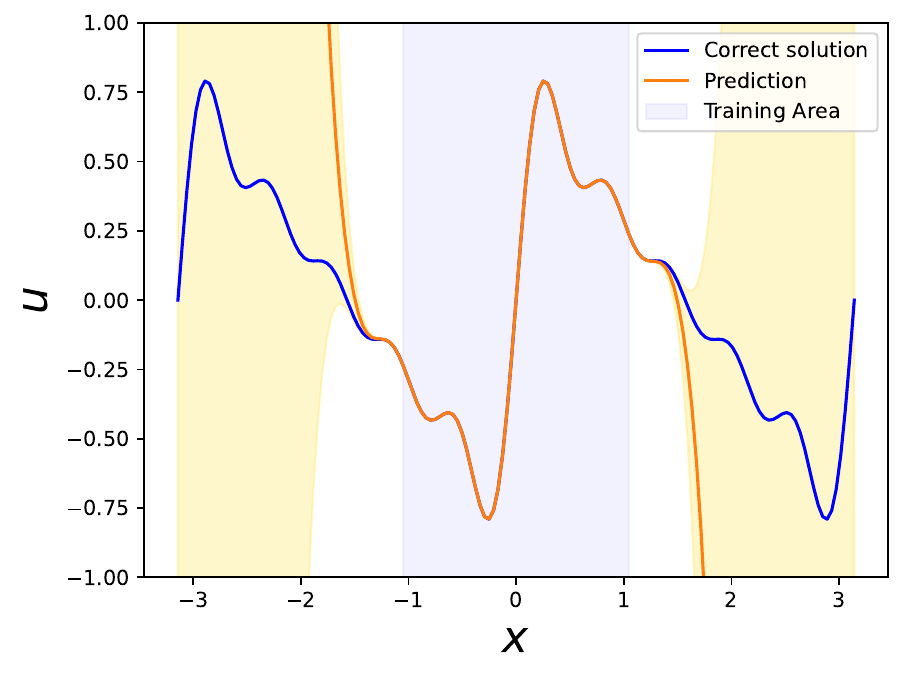}}
    \hfill
  \subfloat[]{%
        \includegraphics[width=0.33\linewidth]{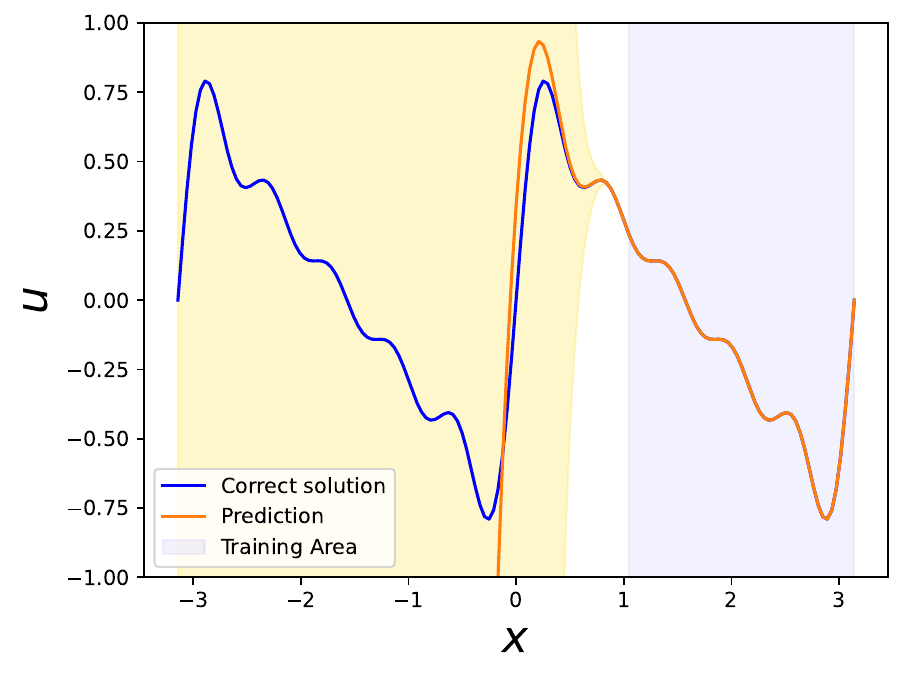}}
    \\
  \subfloat[]{%
        \includegraphics[width=0.33\linewidth]{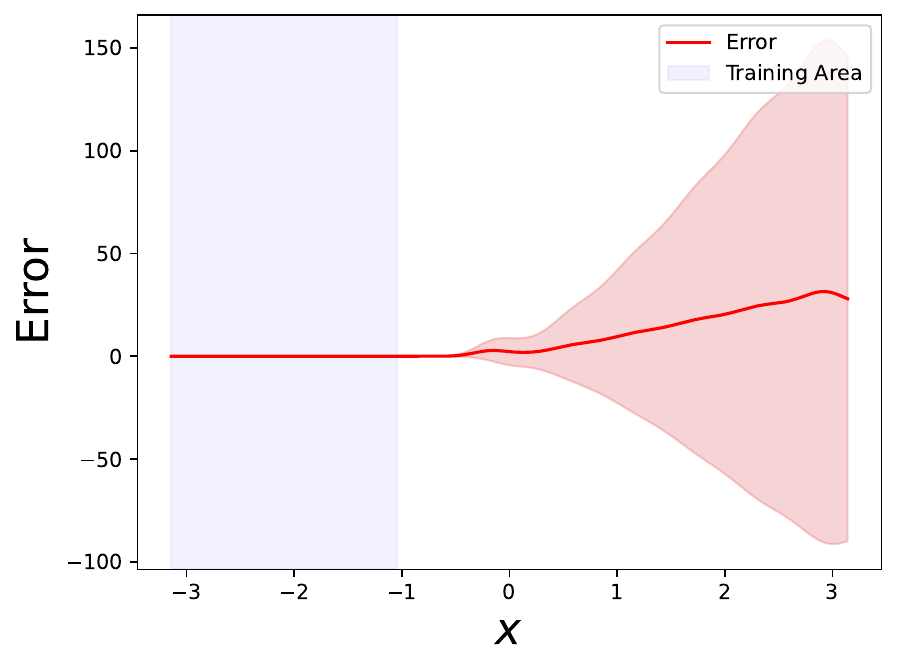}}
    \hfill
  \subfloat[]{%
        \includegraphics[width=0.33\linewidth]{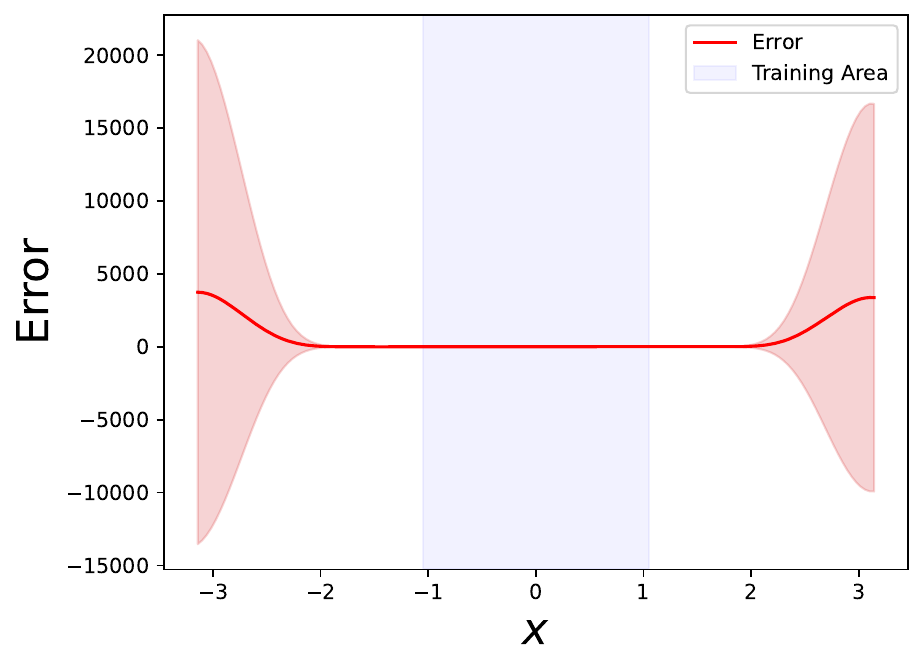}}
    \hfill
  \subfloat[]{%
        \includegraphics[width=0.33\linewidth]{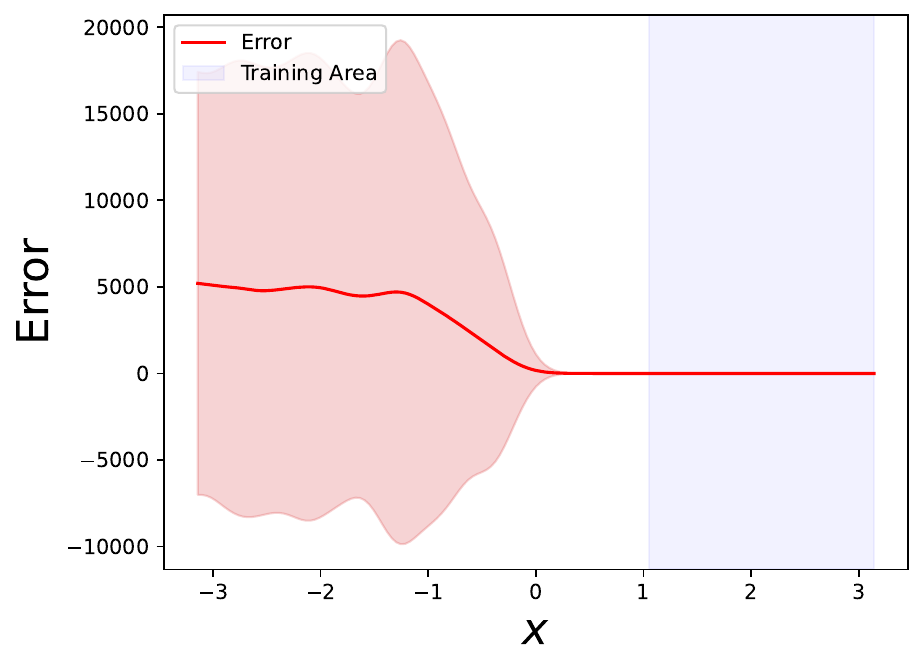}}
        
  \caption{Prediction and ground truth (a,b,c) and related error (d,e,f) for PINNs trained in the respectively light blue area. The shaded gold and red areas represents two times the standard deviation of the prediction and the error of PINNs respectively.}
  \label{PINNS_in_three_domains}
\end{figure}

We now proceed to investigate the concept of generalization for a PINN architecture expressed in Section \ref{sec:100_Method}. Instead of solving Equation \eqref{poisson} in the whole domain, we restrict the training to subsets, and showcase the behavior of the predicted solution outside the training domain. We first split the training domain into three subdomains, namely $\Omega_1 = [-\pi, -\frac{\pi}{3}]$, $\Omega_2 = [-\frac{\pi}{3}, \frac{\pi}{3}]$, and $\Omega_3 = [\frac{\pi}{3}, \pi]$ and perform training of 100 randomly initialized PINN architectures with identical structures, in order to grasp the effect of stochasticity in the training procedure. From an analytical perspective, the exact solutions $u^j$ solving Equation \eqref{poisson} in the respective subdomains $\Omega_j$ are all represented by the same function, due to the uniqueness of the solution. This would mean that each $u_\theta ^j$ in each $\Omega_j$ should behave the same in the whole $\Omega$, which is not the case in practice. Based on the continuity of $u_\theta$ and the target function, the only fair assumption is that the PINN prediction will shortly follow the analytic function. The results plotted in Figure \ref{PINNS_in_three_domains} show the behavior described above, along with the expected degradation of prediction outside the training area.

As expected, predictions are extremely accurate in the training area and maintain precision close to its boundaries. Indeed, since the network is able to properly approximate the derivatives of the target function at some point, then it can also approximate the target function properly in a ball around said point. However, the prediction quickly becomes unreliable and subject to stochasticity further away from the training domain. The average PINN prediction suggests that an accurate function evaluation can be done with confidence only in the immediate proximity of the convex hull defined by the collocation points used in training. This concept is aligned with the analytical error bounds in \cite{mishra2022estimates, pinn4navierstokesbounds}, where the tightest error estimates present an exponential increase in time when moving far away from the training area. However, this does not strictly imply that generalization is not possible, which is why it is important to understand whether the behavior of the PINN outside the training domain can be controlled through the hyperparameters set before the training process.

\subsection{Statistical Analysis}

In this section,  we provide a preliminary analysis of the potential of PINN to generalize to points outside the training domain. In particular, we first want to evaluate whether generalization to points outside the training domain is possible at all for a trained PINN architecture. To do so, we test an empirical bound on the probability of a PINN being able to achieve a prescribed precision in some area of size $\delta$ outside of the training domain. In order to capture more variability, we study the value of the generalization level obtained before the minimization over $u_\theta$ of Definition \ref{generalization_metric}.

\begin{equation}\label{generalization_metric_step_one}
    g^\epsilon _l(u_\theta ) =  \max_{\Omega_G \subseteq \Omega \setminus \Omega_T} \bigl\{ \frac{l(\Omega_G)}{l(\Omega_T)} \quad \text{s.t.:} \\ || u(x) - u_\theta (x) || \leq \epsilon \quad \forall x\in \Omega_G  \bigr\}
\end{equation}

\begin{figure} 
    \centering
  \subfloat[]{%
       \includegraphics[width=0.5\linewidth]{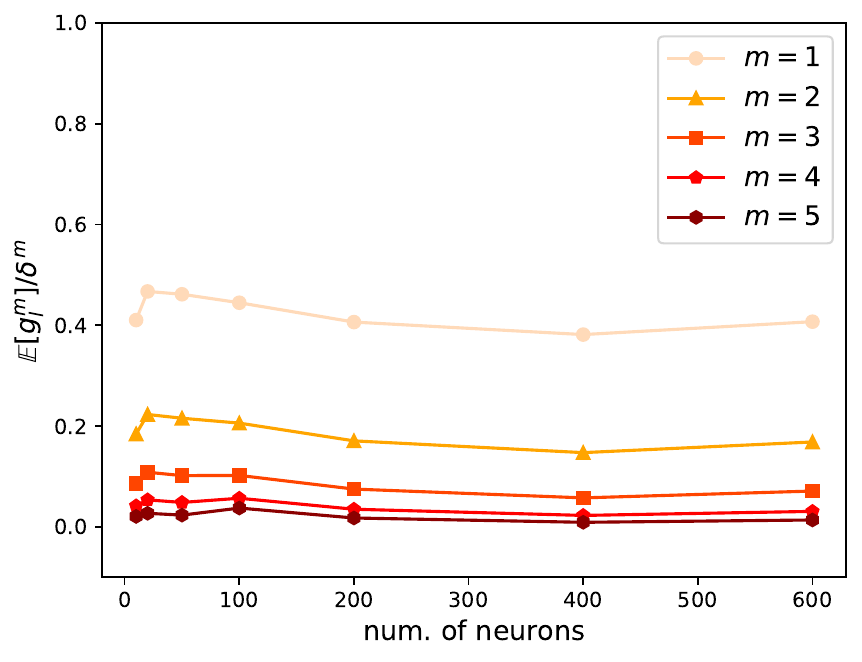}}
    \hfill
  \subfloat[]{%
        \includegraphics[width=0.5\linewidth]{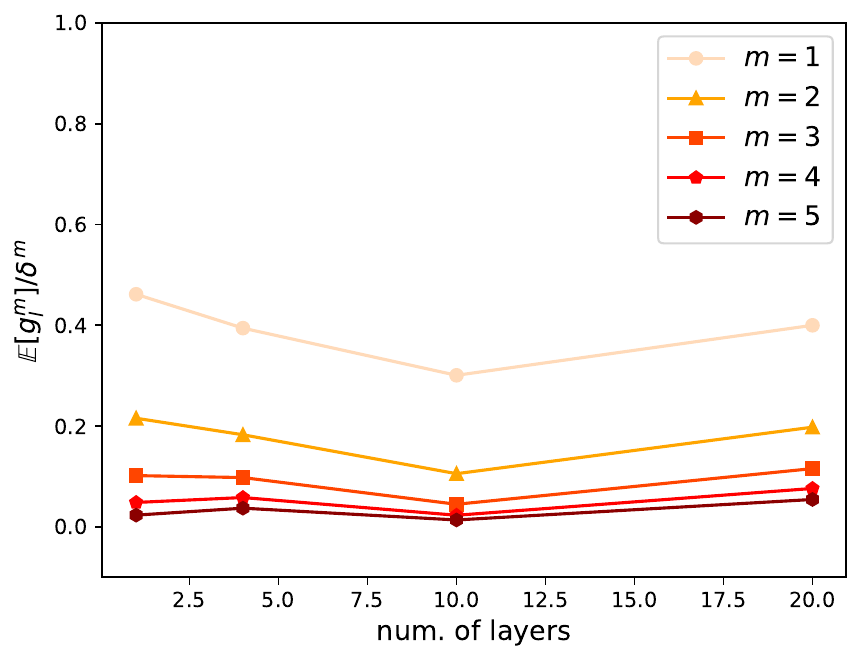}}
    \\
  \subfloat[]{%
        \includegraphics[width=0.5\linewidth]{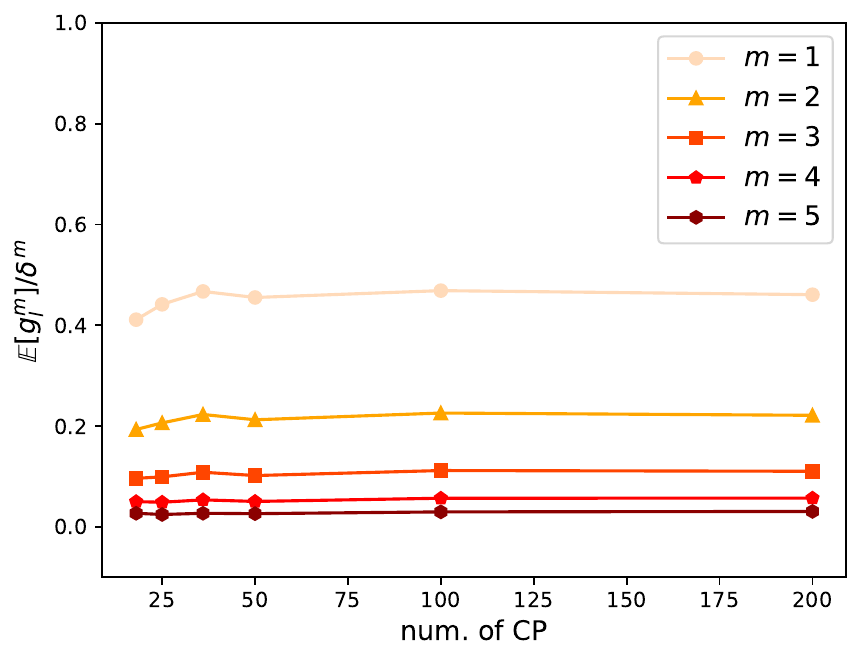}}
    \hfill
  \subfloat[]{%
        \includegraphics[width=0.5\linewidth]{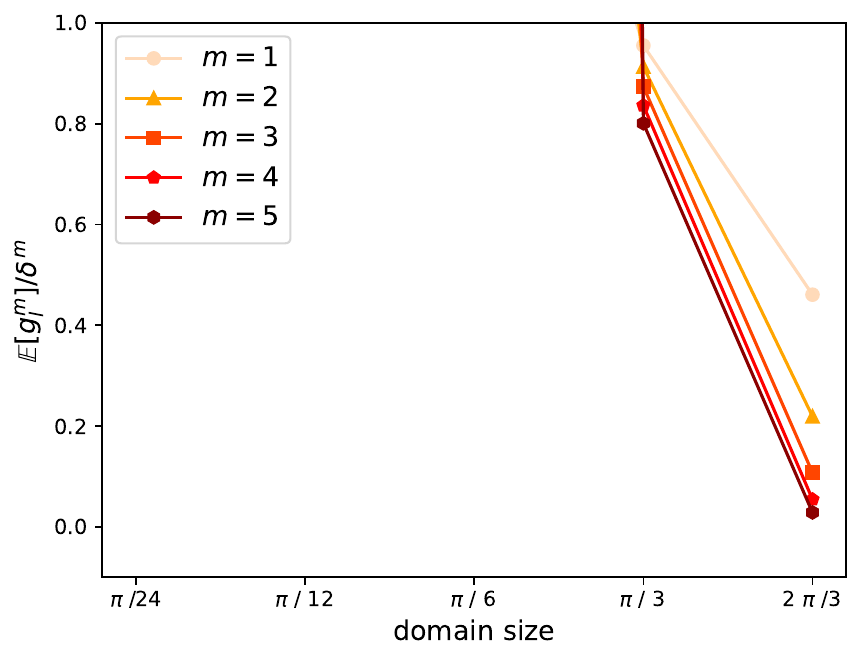}}   
  \caption{Bounds on $P( g_l \geq \delta )$ for different number of neurons (a), layers (b), number of collocation points (c) and domain size (d) for $\delta = 0.25$. Values are limited between 0 and 1, no values indicates a bound larger than 1.}
  \label{Markov0.25}
\end{figure}

Hence, for every neural network and every architectural and algorithmic setup, we consider the variable $g^\epsilon _l$, in Definition \ref{generalization_metric_step_one}, as the subject of our statistical analysis. Then, we study the probability $P( g^\epsilon _l \geq \delta )$ with $\delta > 0$. The value $g^\epsilon _l$ represents a length, which means it is always non-negative. Therefore, we can bound the probability $P(g^\epsilon _l > \delta)$ with Markov's inequality. However, Markov's inequality typically represents a very lose probability bound, which is why one can consider the version for higher moments of order $m$, since the bound holds for any value of $m$. The general formulation is given in Equation \ref{Markov}, the classical Markov's inequality is obtained when $m=1$. Results of the obtained bounds are available in Figure \ref{Markov0.25} for $m=1,\dots,5$, $\delta = 0.25$ and $\epsilon = 10^{-3}$.

\begin{equation}\label{Markov}
    P(|g_l| \geq \delta) \leq \frac{\mathbb{E}[|g_l|^m]}{\delta^m}
\end{equation}

The reasoning behind the choice of Markov's inequality is that for $m$ large enough, the bound will converge to zero if there is no evidence that the generalization level could be greater than $\delta$. Indeed, this is what happens for the case of $\delta = 0.25$ except for small domain sizes, where no relevant bound can be given through Markov's inequality. This suggests that, in general, PINNs fail to extrapolate on domains 25\% larger than the training area. On the other hand, for values of $\delta$ around $0.1$ the obtained bound becomes almost always greater than $1$, which does not provide any relevant insight.

Assessed that PINNs can generalize to a relatively small region outside the training area, we proceed to identify whether the network hyperparameters have a statistically significant impact on the generalization potential by performing the Kruskal-Wallis H test for independent samples \cite{KWHtest}. The latter sets as the null hypothesis the equality in the median of samples coming from multiple groups having independent probability distribution functions. The Kruskal-Wallis H test is based on the rank of the data samples and rejects the null hypothesis when one of the samples appears to have greater values than the other. For our analysis, the chosen groups are identified by $g^\epsilon _l$ values obtained by varying one of the aforementioned hyperparameters.

\begin{table}[h!]
\setlength{\tabcolsep}{18pt}
\centering
\begin{tabular}{|| c || r | r | r | r ||} 
 \hline
 Hyperparameter vs $\epsilon$  & $1e-2$ & $1e-3$ & $1e-4$ & $1e-5$ \\ [0.3ex] 
 \hline\hline
 N. of Neurons & 238.15 & 183.46 & 206.35 & 235.19  \\
  \hline
 N. of Layers & 37.09 & 98.72  & 123.91 & 111.83 \\
  \hline
 N. of CP  & 16.38 & \textbf{6.31} & \textbf{8.86} & 18.53 \\
  \hline
 Domain Size & 48.68 & 260.27  & 324.89 & 346.30 \\
 \hline
\end{tabular}
\caption{Kruskall-Wallis' H statistic value for all hyperparameters and all values of $\epsilon$ studied. Values in bold refer to cases where statistically significant differences can not be validated}
\label{table:1}
\end{table}

The test results are presented in Table \ref{table:1}, where the value of the H statistics are shown for each hyperparameter-based group and each value of $\epsilon$. In the aforementioned Table, we highlight the H values in bold when the p-value is above $0.01$, which represent cases in which the H statistic can not validate any statistically significant difference in distribution. Except for the number of collocation points used in training, all hyperparameters present a strong statistical significance, for any $\epsilon$, to reject the null hypothesis of the Kruskall-Wallis test.
This means that the values of $g^\epsilon _l$ are directly influenced -- with very high probability -- by each of the hyperparameters considered in our study. To identify where the statistically significant differences appear within each group, we also perform pairwise Mann-Whitney U tests -- the two-sample version of the Kruskall-Wallis test --  for each of the hyperparameters that have shown statistically significant differences. The results obtained through these post-hoc tests can be summarized as follows: 

%



\begin{itemize}
    \item \textit{Number of Neurons}: statistically significant differences seem to be present in blocks. Networks with 10 to 100 neurons appear to present similarities within each other, as do networks with 200 to 600 neurons. However, the null hypothesis is always rejected between the two groups, regardless of the values of $\epsilon$.
    \item \textit{Number of Layers}: similar to the case of the number of neurons, results obtained for architectures with 4 or more layers appear to be correlated, while the null hypothesis is always rejected between the one-layered architecture and the other networks.
    \item \textit{Domain Size}: The null hypothesis is always rejected, except for the two configurations with larger domain size, which seem to be correlated when $\epsilon$ is very small.
\end{itemize}

Finally, we can conclude that varying the number of collocation points does not significantly influence the generalization level of a PINN. On the other hand, the network complexity and the size of the PDE domain have a statistically significant impact on the value of the generalization level.\\

%% file: Sections/200_Results.tex
In this section, we provide numerical results on the power of generalization for the aforementioned hyperparameters of the PINN algorithm.  The results obtained for the $G_l$ metric are presented alongside training time information, to provide a more complete understanding of the benefits of specific parametric settings. Moreover, for each hyperparameter, we also provide further visualizations of the behavior of the predictions when the related hyperparameter changes.  For all hyperparameter analysis, we perform training of 100 randomly initialized PINN architectures restricted to $10^4$ Adam iterations with learning rate $10^{-3}$ and a subsequent L-BFGS optimizer limited to $10^4$ iterations and $10^{-8}$ tolerance. The results are then throughly discussed in Section \ref{sec:300_Discussion}.

\subsection{Network Complexity}\label{sec:gen_network_complexity}

The first case taken into consideration is the complexity of the network adopted to define the PINN architecture. The analysis is divided into the number of neurons and the number of layers, to properly grasp the effect of each of the two components.

\subsubsection{Number of Neurons}

\begin{figure} 
    \centering
  \subfloat[]{%
       \includegraphics[width=0.48\linewidth]{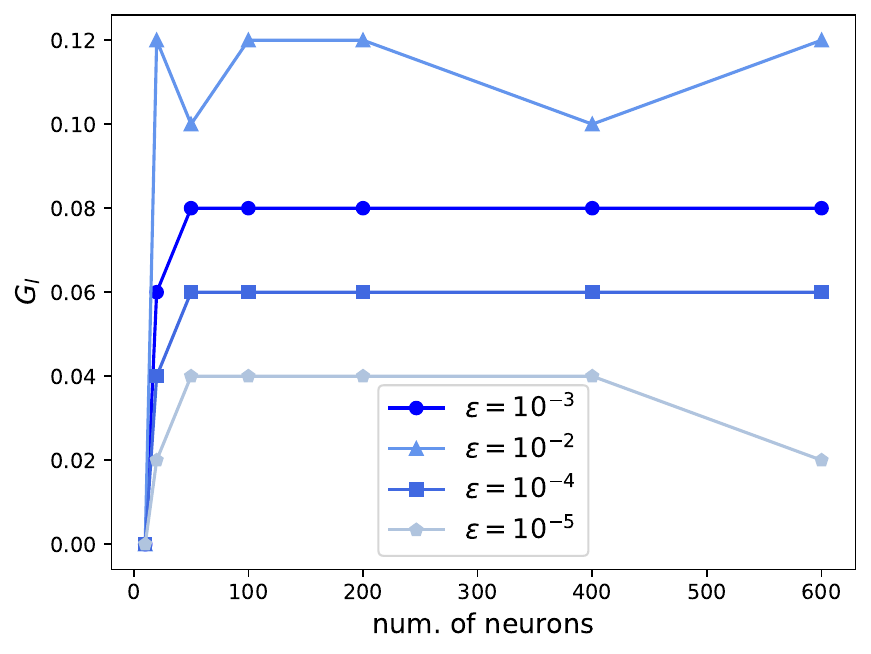}}
    \quad
  \subfloat[]{%
        \includegraphics[width=0.48\linewidth]{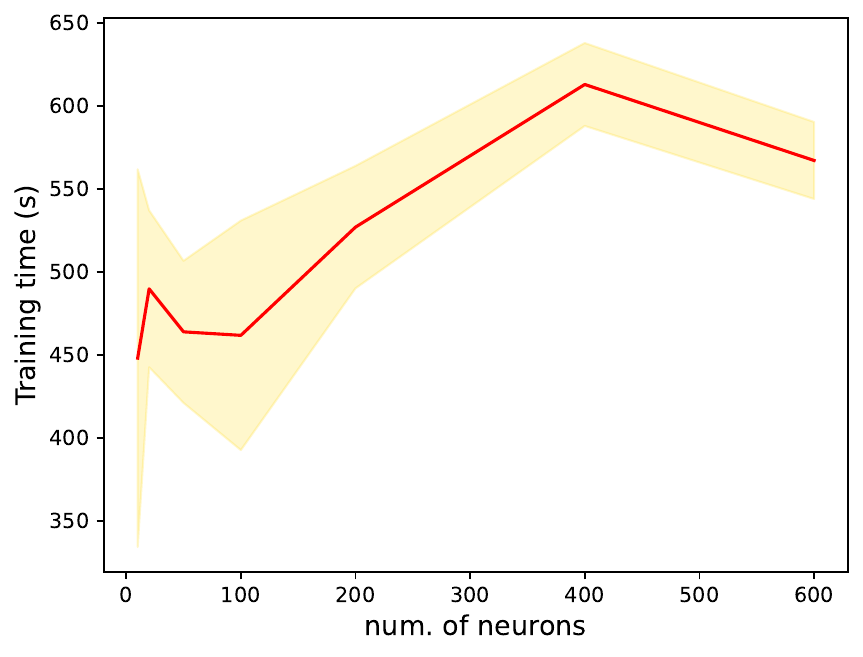}}

  \caption{$G_l$ values for different values of $\epsilon$ when the number of neurons is increased (a) and training time (mean and variance) of each training configuration (b).}
  \label{number_of_neurons_MG} 
\end{figure}

We consider the number of neurons of single-layered neural network architectures. We test several PINN architectures characterized by the number of neurons of their hidden layer and evaluate them based on the value of the GL metric. In order to fully grasp the effect of under and over-parametrized regimes, we test architectures with 10, 20, 50, 100, 200, 400, and 600 neurons. The results obtained for this analysis are shown in Figure \ref{number_of_neurons_MG}, along with the average training time per architecture configuration and its variance.

Regarding the value of the generalization metric, the value is always zero for the smallest architecture since, on average, 10 neurons are not enough to learn the target function. Increasing the number of neurons rapidly enhances the generalization power of the architecture, which seems to be saturated as soon as 50 neurons are used. When the number of neurons is further increased, no clear enhancement is obtained in terms of the generalization metric we have defined. The training time, on the other hand, presents a more peculiar behavior. As expected, convergence is faster on average for smaller architectures, but the variability of the training time is instead much larger. The choice of a 50 neurons architecture seems to be the most appealing both in terms of average training time and variability.

\begin{figure} 
    \centering
  \subfloat[]{%
       \includegraphics[width=0.33\linewidth]{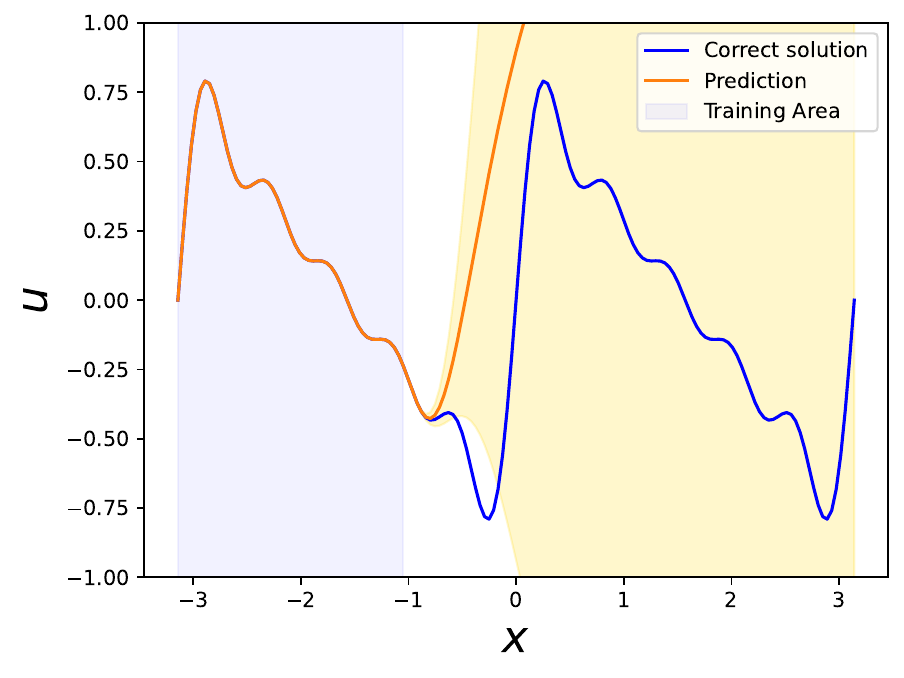}}
    \hfill
  \subfloat[]{%
        \includegraphics[width=0.33\linewidth]{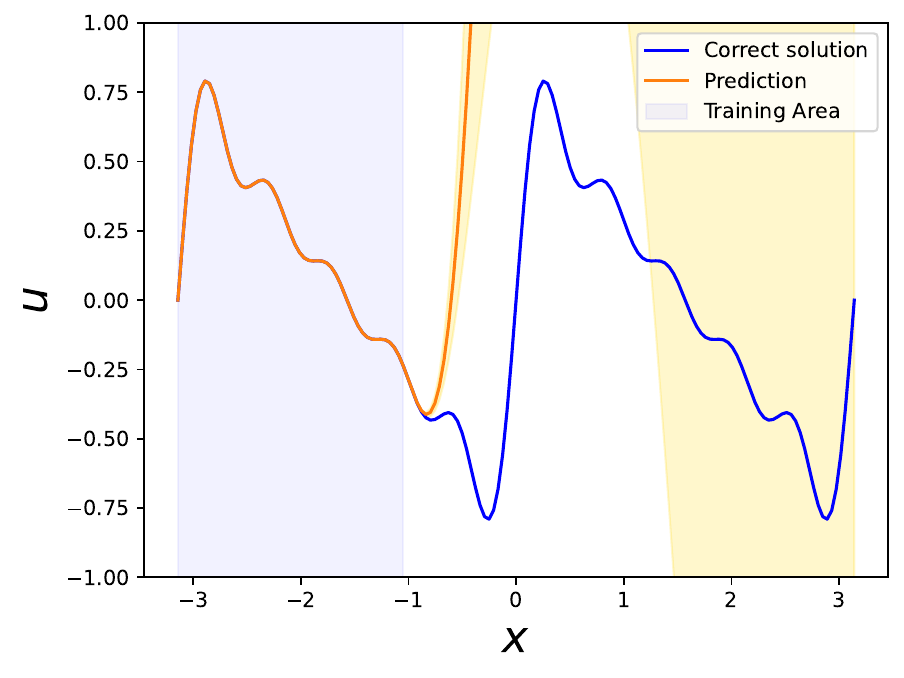}}
    \hfill
  \subfloat[]{%
        \includegraphics[width=0.33\linewidth]{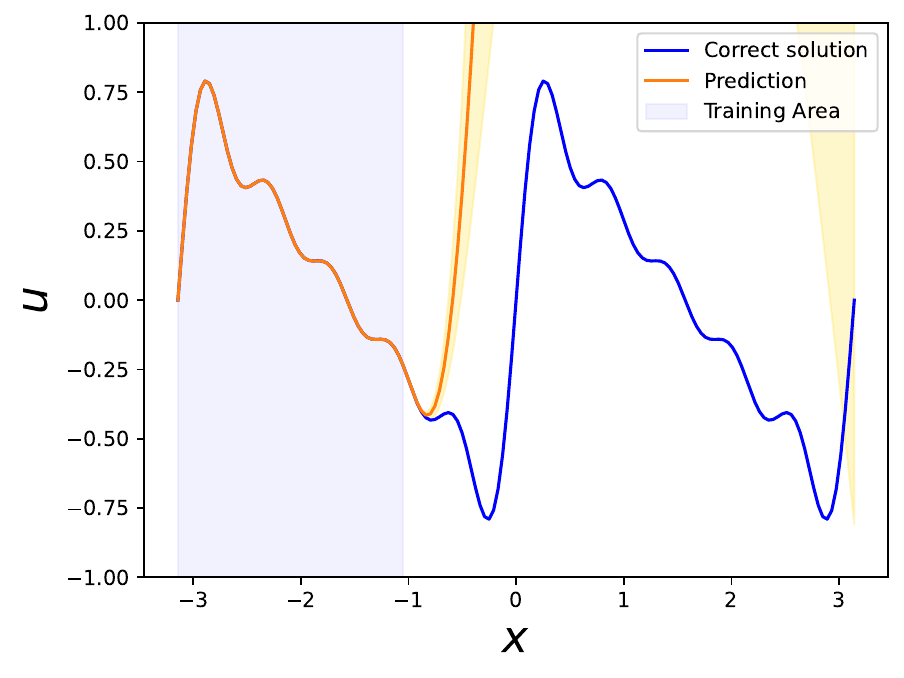}}

  \caption{PINN predictions (with variance) and ground truth for single layer architectures with 20(a), 400(b) and 600(c) neurons, trained in the light blue area.}
  \label{net_complex_showcase} 
\end{figure}

In Figure \ref{net_complex_showcase}, it is possible to also visualize the impact on predictions when the number of neurons in the hidden layer is increased.
Increasing the width of the network seems to generally skew the outside prediction to an overestimation by the PINN, leading to a solution that rapidly departs from the correct one. This effect is not clearly visible for a relatively small number of neurons -- 10, 20, and 50 -- and quickly becomes remarkable when the over-parametrization regime is reached. It is also interesting to notice that the over-parametrization regime does not appear to be beneficial, in terms of the width of the network, to predict the solution of the PDE.

\subsubsection{Number of Layers}

\begin{figure} 
    \centering
  \subfloat[]{%
       \includegraphics[width=0.48\linewidth]{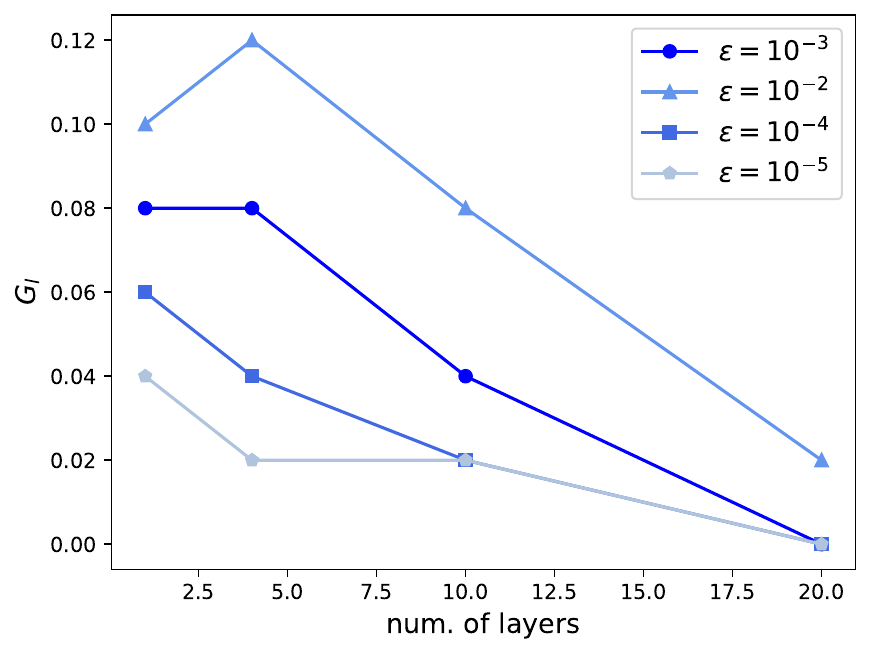}}
    \quad
  \subfloat[]{%
        \includegraphics[width=0.48\linewidth]{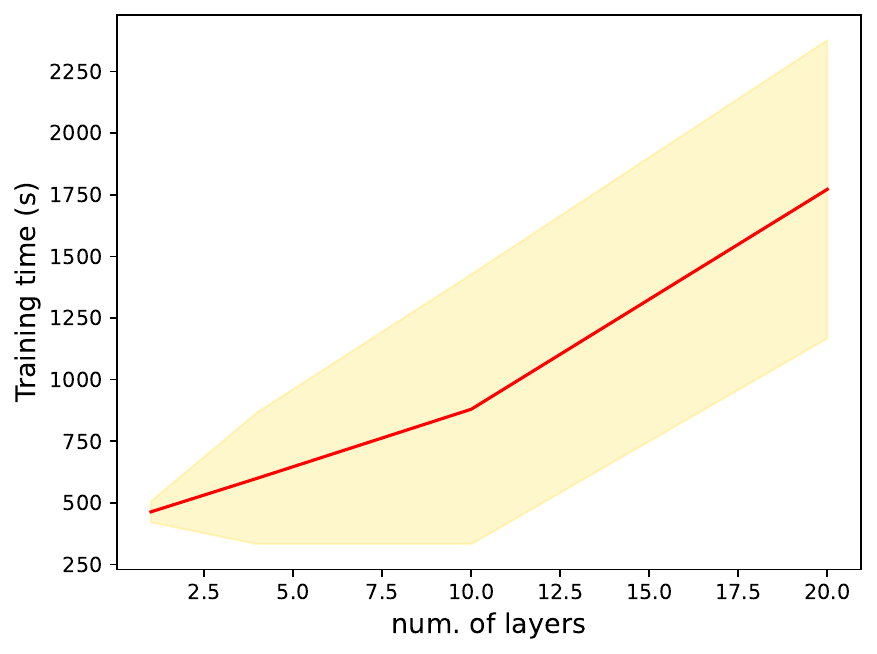}}

  \caption{$G_l$ values for different values of $\epsilon$ when the number of hidden layers of the network is increased (a) and training time (mean and variance) of each training configuration (b).}
  \label{number_of_layers_MG} 
\end{figure}

An analogous analysis is conducted for neural architectures with a fixed number of neurons and a variable number of layers. We provide our analysis with as many as 50 neurons per layer and study architectures with 1, 4, 10, and 20 layers. Once more, the results are presented along with the statistics of the training time per network architecture and can be seen in Figure \ref{number_of_layers_MG}.

The results obtained for the generalization level show on average a decay of the metric value when the number of layers is increased. The exception is the case of five layers or less, where it is possible to notice a slight improvement in the metric for some $\epsilon$. It is also noticeable that the deepest architecture -- 20 layers -- gives almost always zero as generalization metric: this is due to the fact that the training process consists of a limited number of iterations and is stopped afterwards. The number of epochs is not enough to make such a big architecture converge properly. However, we noticed that even when trained for longer periods, such deep architectures struggle to achieve a suitable solution and do not achieve relevant results in terms of the generalization level. Moreover, the results obtained for the training time show an almost linear increase for the average training time, while its variability is significantly smaller when the number of layers is small.

\begin{figure} 
    \centering
  \subfloat[]{%
       \includegraphics[width=0.33\linewidth]{Figures/MultiGenPINN_solution_left_domain_1L_50N.pdf}}
    \hfill
  \subfloat[]{%
        \includegraphics[width=0.33\linewidth]{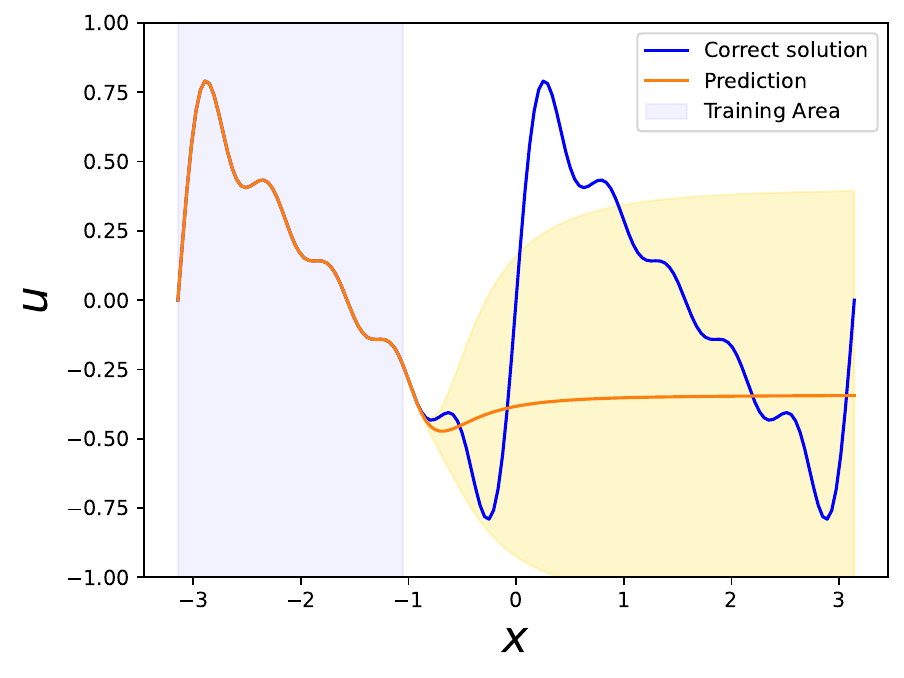}}
    \hfill
  \subfloat[]{%
        \includegraphics[width=0.33\linewidth]{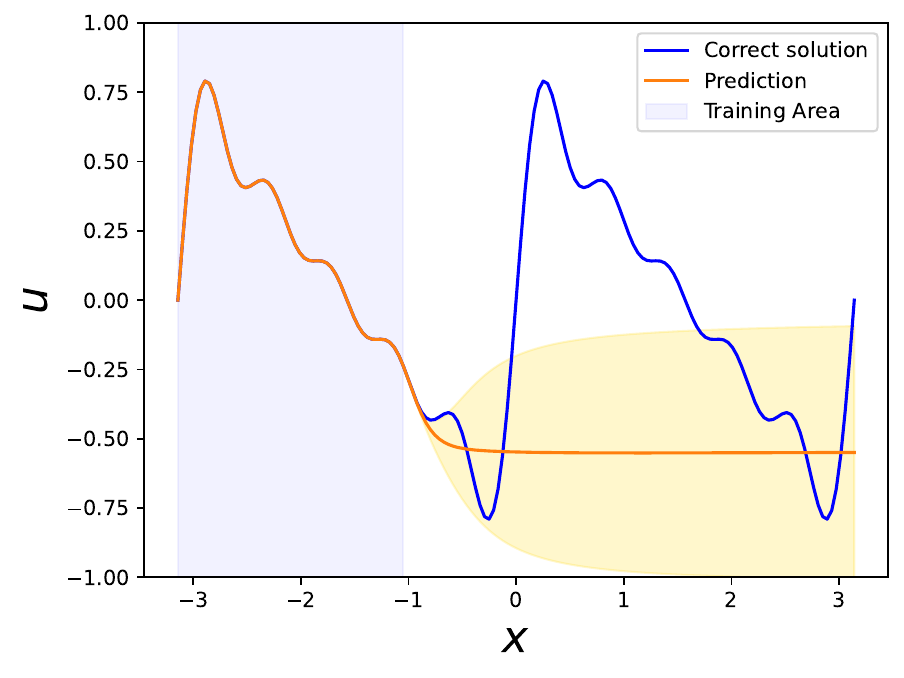}}

  \caption{PINN predictions (with variance) and ground truth for architectures with 1(a), 4(b) and 10(c) layers and 50 neurons per layer, trained in the light blue area.}
  \label{network_depth_showcase} 
\end{figure}

In Figure \ref{network_depth_showcase}, we also provide a visual representation of the effect on the predicted solution of an increasing number of layers. The depth of the PINN appears to smooth out the prediction of the network to a constant value outside the boundary of the training domain. Moreover, this smoothened guess seems to have reduced variability when more layers are added. This impacts the GL metric negatively since its value is based on the outcome of the worst-performing architectures among all the trained ones.

\subsection{Number of collocation points}

\begin{figure} 
    \centering
  \subfloat[]{%
       \includegraphics[width=0.48\linewidth]{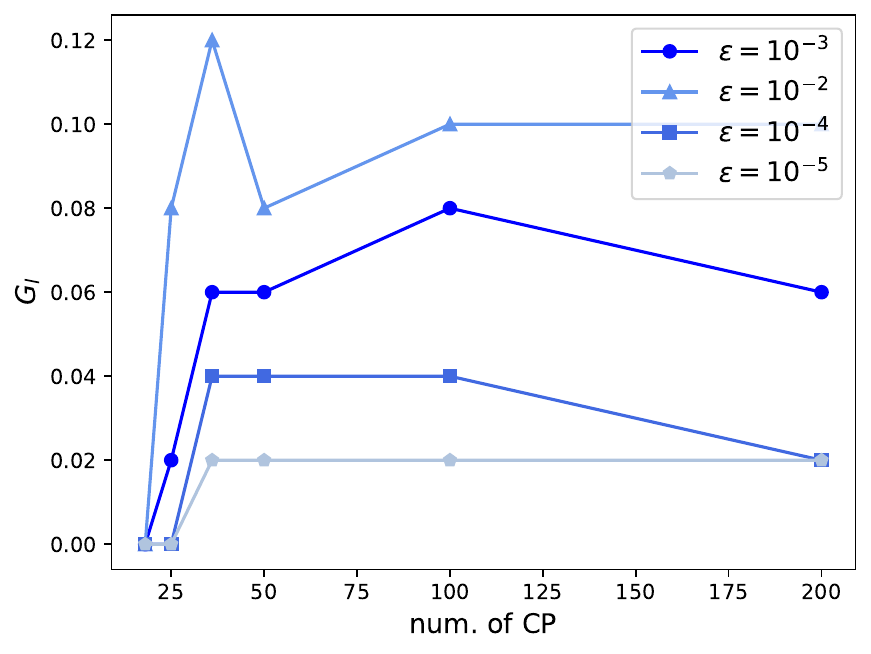}}
    \quad
  \subfloat[]{%
        \includegraphics[width=0.48\linewidth]{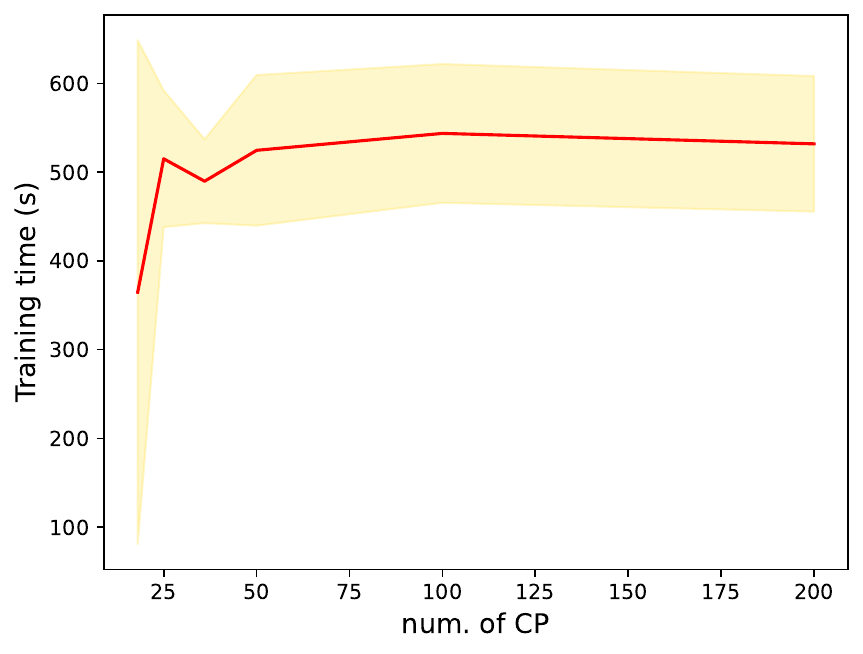}}

  \caption{$G_l$ values for different values of $\epsilon$ when the number of collocation points is increased (a) and training time (mean and variance) of each training configuration (b).}
  \label{NCP_MG} 
\end{figure}

We perform our routine analysis on single-layered neural network architectures with 20 neurons, which take as training input a variable number of collocation points. In our setting we test the case of 18, 25, 36, 50, 100 and 200 collocation points randomly sampled in each iteration with the Latin hypercube sampling strategy over the training domain $[-\pi, -\frac{1}{3} \pi ]$. The results obtained for the generalization level and the training time are given in Figure \ref{NCP_MG}.

The resulting values of our generalization metric show that 18 collocation points are not enough to learn the target function. Once the number of collocation points is increased, the value of the generalization metric also increases. As foreshadowed by the preliminary analysis, the generalization level is not largely affected by the number of collocation points used, as long as enough points are provided. Moreover, for the case used in our analysis, 36 collocation points are sufficient to learn the target function and obtain a good value of generalization. The training time presents a similar steady behavior with respect to the number of collocation points used for training, except for the case when few points are sampled and the training does not reliably converge to the correct solution.

\begin{figure} 
    \centering
  \subfloat[]{%
       \includegraphics[width=0.33\linewidth]{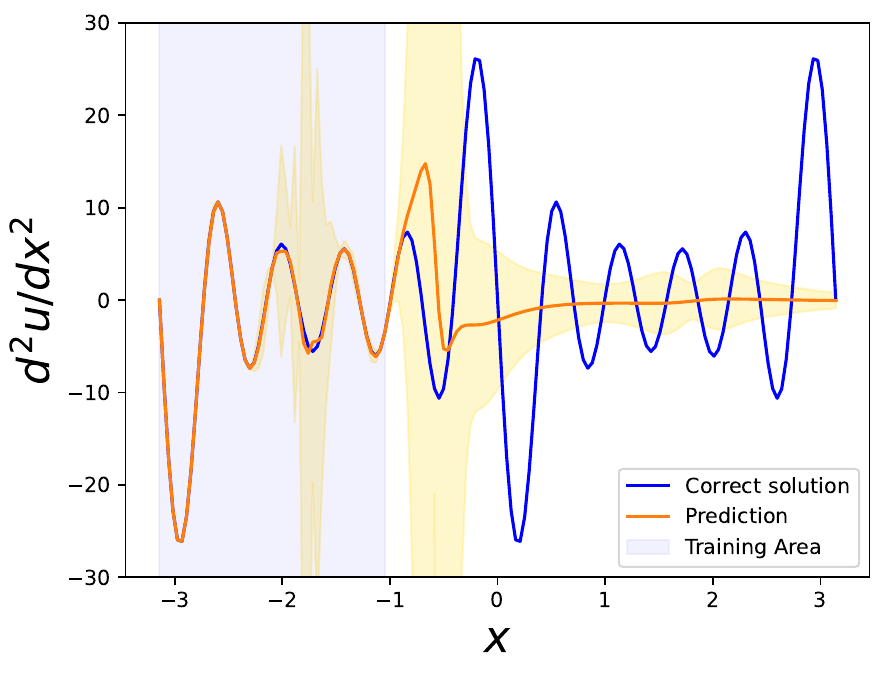}}
    \hfill
  \subfloat[]{%
        \includegraphics[width=0.33\linewidth]{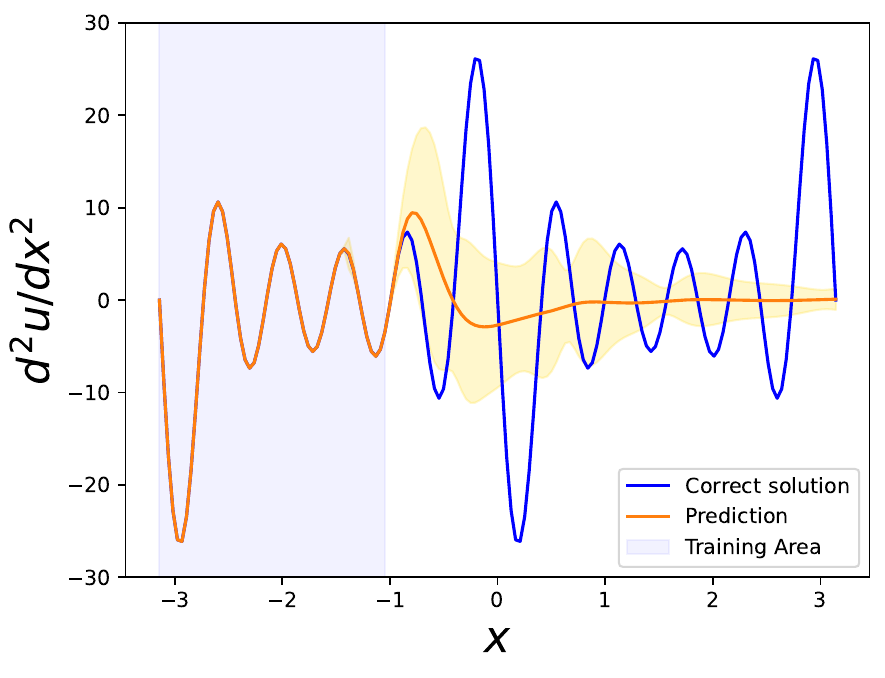}}
    \hfill
  \subfloat[]{%
        \includegraphics[width=0.33\linewidth]{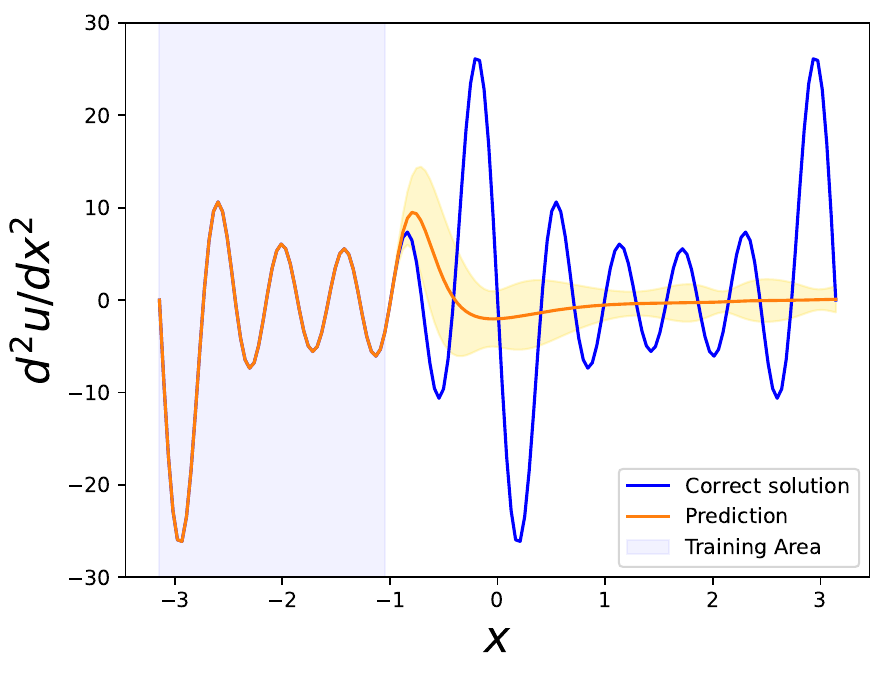}}

  \caption{PINN predictions (with variance) and ground truth of the second derivative of the PDE solution for architectures trained with 18(a), 25(b) and 36(c) collocation points distributed in the light blue area.}
  \label{ncp_showcase} 
\end{figure}

Furthermore, the number of collocation points does not present any major pattern in the prediction of the PDE solution. The only visible effect can be noticed in the prediction of the second derivative of the target function in the low numbers regime. The latter is provided in Figure \ref{ncp_showcase}, which includes the prediction results of the second derivative for PINNs trained with 18, 25, and 36 collocation points sampled in the training domain.
In this visualization, it is possible to see that an increased number of collocation points reduces the variability of the predictions outside the training area. Among those architectures, the latter configuration is the only one that completely reached convergence, as it is possible to identify a non-zero variance in the predictions obtained via performing training with either 18 or 25 collocation points.

\subsection{Domain Size}

\begin{figure} 
    \centering
  \subfloat[]{%
       \includegraphics[width=0.48\linewidth]{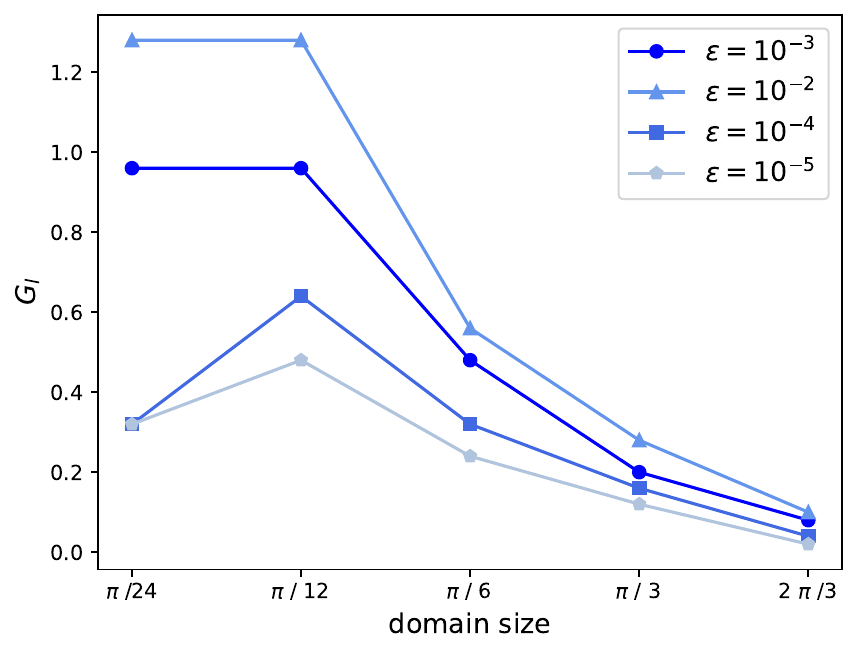}}
    \quad
  \subfloat[]{%
        \includegraphics[width=0.48\linewidth]{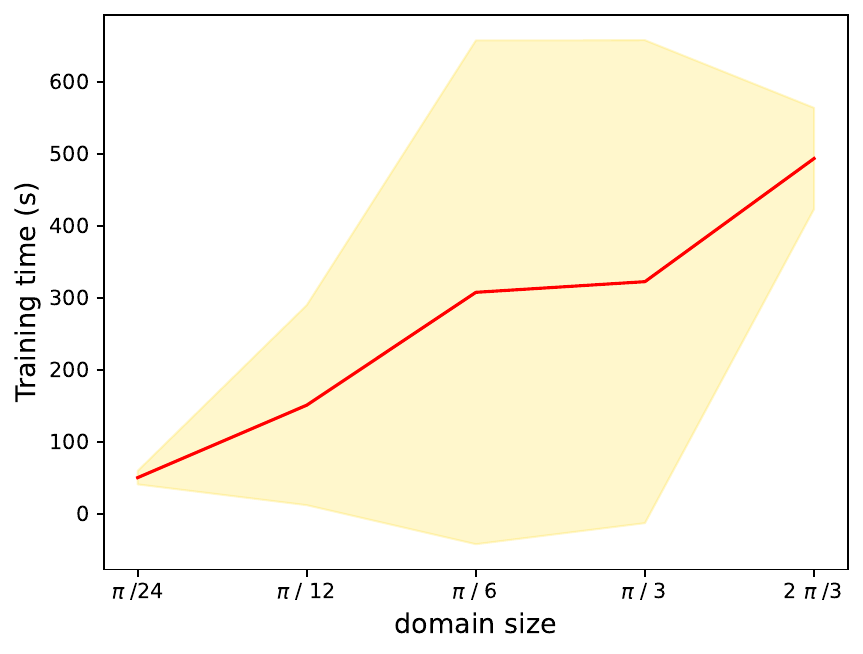}}

  \caption{$G_l$ values for different values of $\epsilon$ when the size of the training domain is increased (a) and training time (mean and variance) of each training configuration (b).}
  \label{domain_size_MG} 
\end{figure}

With domain size, we refer to the length -- or area -- of the region where the collocation points are sampled and where the underlying PDE is defined. For the analysis of this section, we consider decreasingly smaller domains that are defined as $\Omega_i = [-\frac{1}{3}\pi - \frac{2}{3\cdot 2^i }\pi, -\frac{1}{3} \pi ]$ for $i= 0, 1, 2, 3, 4$. Since we have already studied the effect on the generalization of the density of CP in the training domain, we reduce the number of collocation points picked for training the network coherently with the decrease in the size of the domain. The network architecture, instead, is kept fixed to a single-layered neural network with 20 neurons, to properly grasp the effect in generalization of the domain size. For the sake of comparability, the value of the GL metric is computed for the right side of the outer domain, which means that it is computed for all training setups in the area where $x\geq -\frac{1}{3} \pi$. The results for the generalization metric and training time are shown in Figure \ref{domain_size_MG}.

Regarding the generalization level, we can see that architectures trained on the small domains provide a consistent generalization to points outside $\Omega_T$.  Indeed, the GL values obtained in our analysis show a clear decay with respect to the size of the domain. On the other hand, the training time presents an interesting behavior as well: an increased time demand for larger domains on average, but a much larger variability in training time for the intermediate values tested. This is most likely due to the richer variety of functions which exhibits the behavior described by the PDE residuals in the training domain. The case of the smallest domain represents once more an exceptional case, for which the training time presents a substantially small variance.

\begin{figure} 
    \centering
  \subfloat[]{%
       \includegraphics[width=0.33\linewidth]{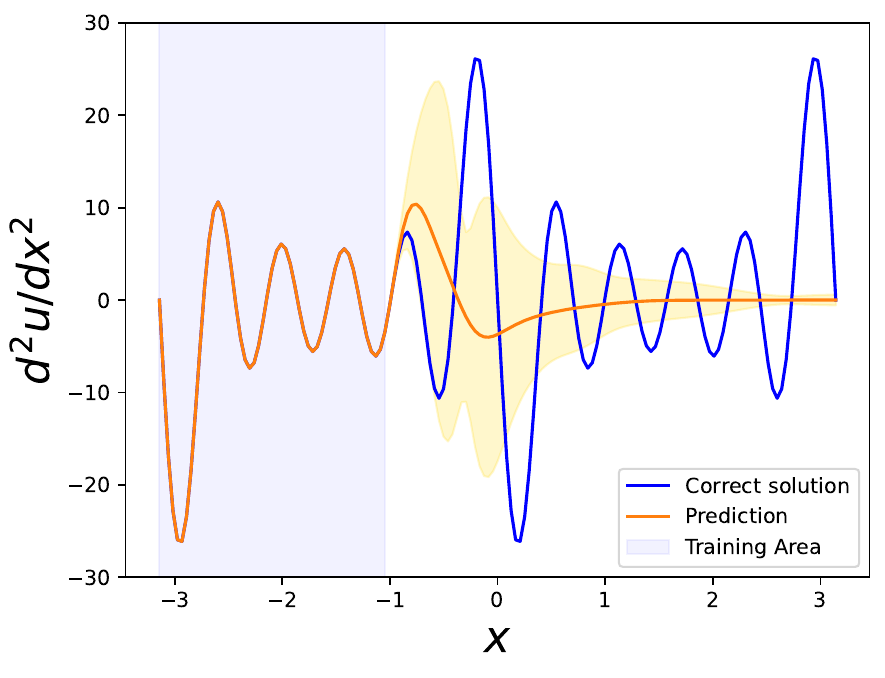}}
    \hfill
  \subfloat[]{%
        \includegraphics[width=0.33\linewidth]{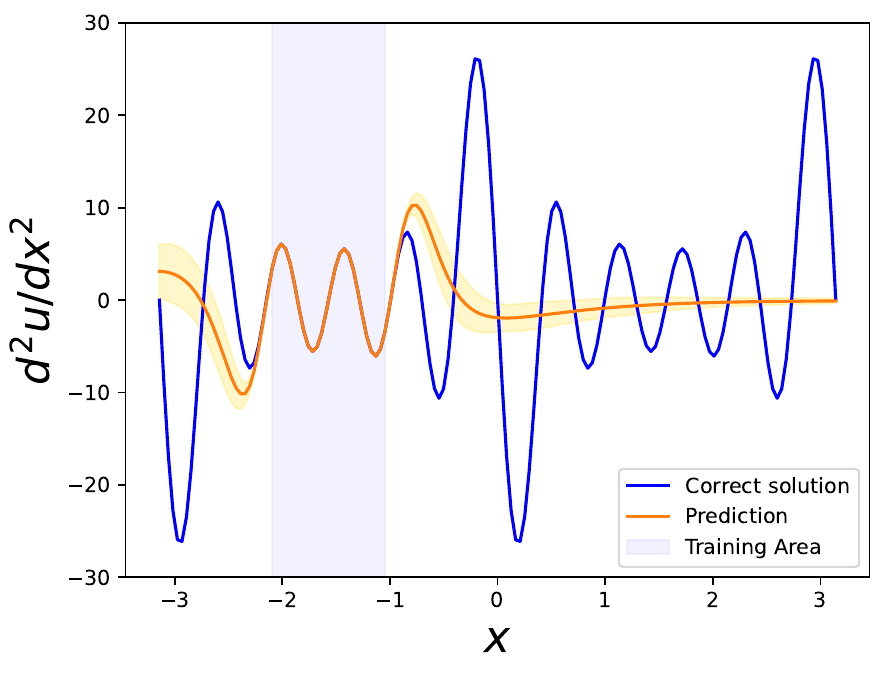}}
    \hfill
  \subfloat[]{%
        \includegraphics[width=0.33\linewidth]{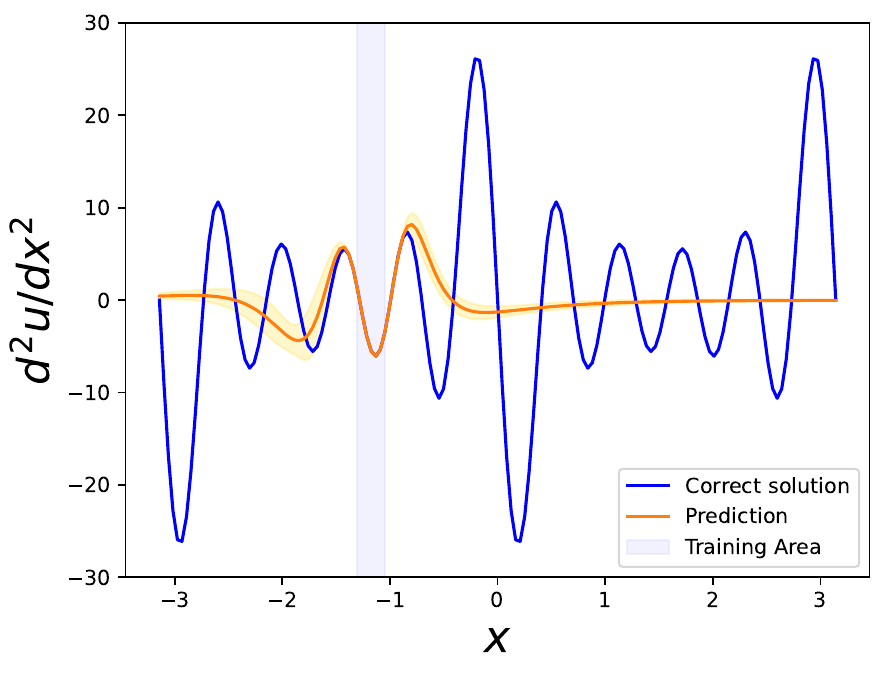}}

  \caption{PINN predictions (with variance) and ground truth of the second derivative of the PDE solution for architectures trained on the domain highlighted by the light blue area.}
  \label{domain_size_showcase} 
\end{figure}

The case of the domain size also represents one of the most interesting impacts on the outcome of the prediction. For visual purposes, we present the effect of the domain size in terms of the prediction of the second derivative of the PDE solution. The results can be seen in Figure \ref{domain_size_showcase} for decreasing domain sizes. It is possible to notice that the main impact on predicting in small domains is given by a reduced variability of the prediction outside of the training area. Since the metric we defined evaluates the best-worst result, the variability of the prediction is a very impactful factor in its value.

%% file: Sections/300_Discussion.tex
In this section, we analyze the results obtained in this paper to provide further insights into the effect of the investigated hyperparameters on the generalization of a PINN. We first briefly comment on the results obtained in our preliminary analysis of Section \ref{sec:X00_Numerics} and then focus on the effect on the generalization level and training time of each of the hyperparameters individually. We present the aforementioned results along with additional highlights that were noticed during the training of all the PINN architectures used in this paper and in comparison with the research literature.

Our preliminary analysis highlights how a PINN solution represents an accurate approximation of the target PDE solution $u$, including even its high order derivatives. This property of PINNs is particularly interesting because it represents the most consistent difference between a PDE solution obtained with a classical solver. On the other hand, our analysis clearly shows that a PINN understandably fails to learn the analytical extension of the PDE solution in regions that are far from the collocation points used during training. However, the architecture maintains the accuracy reached inside the training domain when predictions are made in the proximity of its boundaries. This is mainly due to the smoothness of both $u$ and $u_\theta$ and means that a PINN architecture has a potential to approximate the target function even outside the training domain.  Therefore, the outcome of our study is given by which settings should generally be considered good practice when utilizing a Physics-Informed Neural Network architecture.\\

\subsubsection{Network Complexity}

We notice that increasing the width of the network skews the prediction to overestimated results far from the training area. However, the value of the generalization metric is not largely affected by the width of the network. Generally, when training an arbitrary neural network, it is common practice to over-parametrize the latter to be sure that the network can capture the complexity of the target function. This is a recently noticed behavior of neural networks in the overparametrized regime which is regarded in the scientific community as the double descent phenomena \cite{mei2022generalization}. Curiously, the latter is not clearly visible when considering the generalization level introduced in this project. Neural networks are also subject to implicit bias \cite{valle2018implicitbias} which skews them towards learning simple solutions. However, the results obtained so far hint that a smaller architecture might be more suitable for training a PINN as computational time is larger for wide architectures while the generalization level does not largely improve. Moreover, when adopting methods which employs a soft-domain decomposition, largely overestimating the target solution could cause numerical instabilities since the gating mechanism might not ensure that the prediction of a model in a subdomain vanishes far from its domain of evaluation.

Increasing the depth of the network skews the PINN to return a flat prediction outside the training domain, which suggests that no information is somehow gathered from the inner domain. Moreover, the generalization level is worsened by the depth, and training time is visibly increased without additional benefits. However, a flat prediction outside of the training domain could be preferrable for cases of soft-domain decomposition. Moreover, the results obtained hold specifically for the relatively simple PDE considered as a base case. Therefore, the outcome of our analysis does not imply that a smaller network is always a more suitable solution, but rather that it is possible and advantageous -- in terms of generalization level and training time -- to compute relatively simple PDE solutions via a PINN with few trainable parameters, encouraging approaches of domain decomposition.


\subsubsection{Number of collocation points}

The generalization level is not largely affected by the number of collocation points. This implies that if the collocation points sampled are enough to correctly predict the target function, there is no benefit in adding more, except for cases of adaptive sampling \cite{adaptivesamplingPINNs}. Moreover, it is also worth noting the unexpected increase in the variability of training time when few collocation points are given. This is most likely related to the fact that using few points generates several local minima in the loss landscapes which are seen as acceptable to the algorithm. However, the lack of impact in training time is most likely because the computational overhead created by just 200 collocation points is negligible for a use case like the one we consider, for which the computation of the residuals is extremely fast. Nevertheless, this is not always the case, especially for multidimensional and highly non-linear equations such as Navier-Stokes'.\\

\subsubsection{Domain size}

In terms of generalization level, we notice that training on a small domain appears to be beneficial for the PINN architecture. Indeed, the generalization level decays when the size of the training area increases In particular, the numerical value of the generalization level obtained for small domain sizes is also rather surprising. Indeed, the values obtained indicate that the longest segment in which all architectures generalize well can be just as big as the domain used for training -- on the right side --. This means that, by training in the selected domain, we get the correct solution in a domain twice the size.

In general, training a PINN architecture is slow and difficult. However, as for the case of a few parameters, we noticed that the PINN training is at times incredibly fast, comparable with the case of the smallest domain size considered in our analysis. The results obtained are coherent with the available literature: indeed, PINNs are known to struggle when learning functions defined in big domains \cite{causalityisallyouneed, wang2022and} and their training can be eased by learning on collocation points which are given sequentially to the algorithm \cite{mishra2021enhancing, krishnapriyan2021characterizing} or selected during training through ensemble agreement \cite{haitsiukevich2023improvedensemble}. In fact, learning on smaller domains requires a surprisingly short training time which also presents an extremely low variability. Since the architecture used is unchanged and the generalization metric actually improves, this could be due to the fact that a smaller domain acts as a regularization term, which skews the learning toward a specific solution in the search space of the neural network. This can also be explained through the study proposed in \cite{rohrhofer2022understanding}, where the authors noticed that learning on small domains reduces the number of fixed points in the loss landscapes, which carries substantial benefits as better iterative training speed and precision.

\subsection{Extension to different cases and higher dimensions}

The results obtained in this paper can be extended to any kind of partial differential equations. In particular, if the considered PDE is monodimensional, it is rather straightforward to repeat the tests performed in our analysis. On the other hand, extending our analysis to PDEs in higher dimensions can be non-trivial conceptually and, more specifically, computationally.

Conceptually speaking, the proposed generalization level $G_l$ represents the value of some distance $d$ of data points from the convex hull of the training samples. When new unseen samples are fed to the model, the predicted value is guaranteed to be precise -- up to a suitable threshold $\epsilon$ --, provided that the new samples are at most $G_l$ away from the training data. In machine learning applications, it is fundamentally hard to guarantee a certain precision, as the underlying problems are typically affected by stochasticity. Thus, any guarantees are typically possible only with high probability and are not necessarily true in general. However, stochasticity should not appear when the underlying problem is the solution of a PDE, which is why we adopted a definition for $G_l$ that should minimize the stochasticity intrinsic in a neural network architecture.

Based on the considerations above, it is possible that for different settings it is better to consider an alternative formulation of the generalization level which does not focus on the maximal set where a suitable generalization is achieved, but rather on the distance from the training set within which every prediction remains accurate. From the perspective of Definition \ref{generalization_metric}, instead of searching for the maximal set $\Omega_G \subseteq \Omega\subseteq \Omega_T$, the aforementioned alternative should focus on a maximal distance value $d_M$, computed through some point-to-set distance $d$ such as the Wasserstein distance. Such a metric can be defined as in Definition \ref{generalization_metric_alternative}.

\begin{equation}\label{generalization_metric_alternative}
    G^\epsilon _{l, \text{alt}} (\Theta, H ) = \min_{u_\theta} \Bigl\{ \max_{d_M} \bigl\{ d_M \quad \text{s.t.:}  \\ \forall x \quad \frac{d(x, \Omega_T )}{l(\Omega_T)} \leq d_M \implies || u(x) - u_\theta (x) || \leq \epsilon  \bigr\} \Bigr\}
\end{equation}

Using either formulation of $G_l$ is equivalent when the underlying problem is monodimensional. However, for higher dimensions, the two metrics can return many different results as Definition \ref{generalization_metric} allows for arbitrary complex shapes of $\Omega_G$, whose area represents the metric value, while Definition \ref{generalization_metric_alternative} substantially limits the shape of $\Omega_G$ to that of a ball around $\Omega_T$. However, the latter formulation represents a computational advantage over the former since it does not require approximating the area of an arbitrarily complex shape.

%% file: Sections/X_Conclusion.tex
The initial goal of this paper was to shed light on the behaviour of the prediction of a PINN outside of its training domain. Through an academic example, we showcased the potential and limits of PINNs in generalization and, afterward, we analyse whether some experimental settings can naturally influence it. The PINN architecture has shown to have a limited potential in generalization as it generally fails to provide the analytical extension of $u$ in an area 25\% bigger than the training domain for a relitively simple 1D problem. However, we have seen that it has the potential to return consistent prediction if evaluated in an area 10\% larger than the training domain. This region might seem negligible, but it represent a consistent size considering that no data was given to the network during training for that area. Hereinafter, by considering the training time, the effect on the prediction, and our metric, we obtained insights into the influence of training hyperparameters on the -- limited -- potential in generalization of PINNs and, most importantly, into the behaviour of their prediction close to the training region.

-

The analysis performed with our metric showed counterintuitive results that are aligned with the scarce theoretical background available for PINN architectures. The majority of the hyperparameters studied have shown to impact the parametrization learned by the PINN training, either in the generalization level or in training time. Despite their effects being in general relatively small, the changes in the neural network's output are often remarkable. For instance, based on the results shown in this paper, the recently worshipped concept of over-parametrization of neural networks might not be the most efficient choice when applying PINNs. Indeed, our analysis highlighted a strong overestimation for wide architectures, a flatter prediction for deep ones, and a consistent extrapolation for networks trained on small areas. The latter facts underline the perks and risks that can occur when combining domain decomposition techniques with PINNs. The same holds for methods that relies on ensemble agreement for training as in \cite{haitsiukevich2023improvedensemble}.

Regarding the usage of the generalization level, future work should provide further results in terms of the generalization level for additional one-dimensional cases, such as a transport and/or advection-diffusion equation. The goal of performing such analysis would be to further validate the results obtained in this paper and explore additional insight into the training of PINNs and the behavior of its prediction outside the training area based on the kind of differential equation that should be solved. An additional interesting research direction is given by the extension of the generalization level $G_l$ to higher dimensions, to enable the application of the metric to more general non-physics-informed architectures and possibly connecting it to existing metrics for the generalization error of arbitrary models.

Based on the results obtained in this paper, future research on the PINN architecture should be skewed in directions that recall divide-and-conquer approaches as the one described in \cite{wang2022mosaic}. Indeed, the most salient factor noticed in our experiments is given by the advantage of training PINNs in small regions. Another valuable alternative is represented by methods that consider sequential training as intrinsic to the architecture, as in \cite{causalityisallyouneed}. Such extensions of PINNs should be accompanied by additional research on optimal neural network design and training setup. Another interesting research direction could be given by adopting the proposed metric to optimize ensembles training methods, which have been proven effective for Physics-Informed models in \cite{haitsiukevich2023improvedensemble}.

\section*{Acknowledgements}
We acknowledge the financial support of BMW AG, Digital Campus Munich (DCM), through the ProMotion program. This project was also funded by the Basque Government: KK2020/00049 project through ELKARTEK program, and the BERC 2022-2025 program. The Spanish Ministry of Science, Innovation and Universities is also acknowledged for its support through project PID2019-104966GB-I00 and through BCAM Severo Ochoa excellence accreditation CEX2021-001142-S/MICIN/AEI/10.13039/501100011033.

